\documentclass[runningheads]{llncs}

\usepackage{eccv}

\usepackage{eccvabbrv}
\usepackage{url}
\usepackage{listings}
\usepackage{spverbatim}
\usepackage{url}
\usepackage[utf8]{inputenc} % allow utf-8 input
\usepackage[T1]{fontenc}    % use 8-bit T1 fonts
\usepackage{booktabs}       % professional-quality tables
\usepackage{amsfonts}       % blackboard math symbols
\usepackage{nicefrac}       % compact symbols for 1/2, etc.
\usepackage{microtype}      % microtypography
\usepackage{xcolor}         % colors
\usepackage{amsmath}
\usepackage{graphicx}
\usepackage{multirow}
\usepackage{xspace}
\usepackage[accsupp]{axessibility}  % Improves PDF readability for those with disabilities.

\usepackage{hyperref}

\usepackage{orcidlink}
\begin{document}

\title{Gaze-to-text Generation: Beyond Categorical Decoding of Human Attention} 

\titlerunning{Gaze-to-text Generation}

\author{Sounak Mondal\inst{1}\orcidlink{0009-0009-9802-4652} \and
Dimitris Samaras\inst{1}\orcidlink{0000-0002-1373-0294} \and
Gregory Zelinsky\inst{1}\orcidlink{0000-0002-3265-3159
} \and
Minh Hoai\inst{2}\orcidlink{0000-0002-2415-6048
}}

\authorrunning{Sounak Mondal et al.}

\institute{Stony Brook University, Stony Brook, New York, USA 
% \and
% Springer Heidelberg, Tiergartenstr.~17, 69121 Heidelberg, Germany
% \url{http://www.springer.com/gp/computer-science/lncs} 
\and
Australian Institute for Machine Learning, Adelaide University, Australia
% \\
}

\maketitle

\def\mA{\mathcal{A}}
\def\mB{\mathcal{B}}
\def\mC{\mathcal{C}}
\def\mD{\mathcal{D}}
\def\mE{\mathcal{E}}
\def\mF{\mathcal{F}}
\def\mG{\mathcal{G}}
\def\mH{\mathcal{H}}
\def\mI{\mathcal{I}}
\def\mJ{\mathcal{J}}
\def\mK{\mathcal{K}}
\def\mL{\mathcal{L}}
\def\mM{\mathcal{M}}
\def\mN{\mathcal{N}}
\def\mO{\mathcal{O}}
\def\mP{\mathcal{P}}
\def\mQ{\mathcal{Q}}
\def\mR{\mathcal{R}}
\def\mS{\mathcal{S}}
\def\mT{\mathcal{T}}
\def\mU{\mathcal{U}}
\def\mV{\mathcal{V}}
\def\mW{\mathcal{W}}
\def\mX{\mathcal{X}}
\def\mY{\mathcal{Y}}
\def\mZ{\mathcal{Z}} 

\def\bbN{\mathbb{N}} 
\def\bbR{\mathbb{R}} 
\def\bbP{\mathbb{P}} 
\def\bbQ{\mathbb{Q}} 
\def\bbE{\mathbb{E}}

\def\1n{\mathbf{1}_n}
\def\0{\mathbf{0}}
\def\1{\mathbf{1}}

\def\A{{\bf A}}
\def\B{{\bf B}}
\def\C{{\bf C}}
\def\D{{\bf D}}
\def\E{{\bf E}}
\def\F{{\bf F}}
\def\G{{\bf G}}
\def\H{{\bf H}}
\def\I{{\bf I}}
\def\J{{\bf J}}
\def\K{{\bf K}}
\def\L{{\bf L}}
\def\M{{\bf M}}
\def\N{{\bf N}}
\def\O{{\bf O}}
\def\P{{\bf P}}
\def\Q{{\bf Q}}
\def\R{{\bf R}}
\def\S{{\bf S}}
\def\T{{\bf T}}
\def\U{{\bf U}}
\def\V{{\bf V}}
\def\W{{\bf W}}
\def\X{{\bf X}}
\def\Y{{\bf Y}}
\def\Z{{\bf Z}}

\def\a{{\bf a}}
\def\b{{\bf b}}
\def\c{{\bf c}}
\def\d{{\bf d}}
\def\e{{\bf e}}
\def\f{{\bf f}}
\def\g{{\bf g}}
\def\h{{\bf h}}
\def\i{{\bf i}}
\def\j{{\bf j}}
\def\k{{\bf k}}
\def\l{{\bf l}}
\def\m{{\bf m}}
\def\n{{\bf n}}
\def\o{{\bf o}}
\def\p{{\bf p}}
\def\q{{\bf q}}
\def\r{{\bf r}}
\def\s{{\bf s}}
\def\t{{\bf t}}
\def\u{{\bf u}}
\def\v{{\bf v}}
\def\w{{\bf w}}
\def\x{{\bf x}}
\def\y{{\bf y}}
\def\z{{\bf z}}

\def\balpha{\mbox{\boldmath{$\alpha$}}}
\def\bbeta{\mbox{\boldmath{$\beta$}}}
\def\bdelta{\mbox{\boldmath{$\delta$}}}
\def\bgamma{\mbox{\boldmath{$\gamma$}}}
\def\blambda{\mbox{\boldmath{$\lambda$}}}
\def\bsigma{\mbox{\boldmath{$\sigma$}}}
\def\btheta{\mbox{\boldmath{$\theta$}}}
\def\bomega{\mbox{\boldmath{$\omega$}}}
\def\bxi{\mbox{\boldmath{$\xi$}}}
\def\bnu{\mbox{\boldmath{$\nu$}}}                                  
\def\bphi{\mbox{\boldmath{$\phi$}}}
\def\bmu{\mbox{\boldmath{$\mu$}}}

\def\bDelta{\mbox{\boldmath{$\Delta$}}}
\def\bOmega{\mbox{\boldmath{$\Omega$}}}
\def\bPhi{\mbox{\boldmath{$\Phi$}}}
\def\bLambda{\mbox{\boldmath{$\Lambda$}}}
\def\bSigma{\mbox{\boldmath{$\Sigma$}}}
\def\bGamma{\mbox{\boldmath{$\Gamma$}}}
                                  
\newcommand{\myprob}[1]{\mathop{\mathbb{P}}_{#1}}

\newcommand{\myexp}[1]{\mathop{\mathbb{E}}_{#1}}

\newcommand{\mydelta}[1]{1_{#1}}

\newcommand{\myminimum}[1]{\mathop{\textrm{minimum}}_{#1}}
\newcommand{\mymaximum}[1]{\mathop{\textrm{maximum}}_{#1}}    
\newcommand{\mymin}[1]{\mathop{\textrm{minimize}}_{#1}}
\newcommand{\mymax}[1]{\mathop{\textrm{maximize}}_{#1}}
\newcommand{\mymins}[1]{\mathop{\textrm{min.}}_{#1}}
\newcommand{\mymaxs}[1]{\mathop{\textrm{max.}}_{#1}}  
\newcommand{\myargmin}[1]{\mathop{\textrm{argmin}}_{#1}} 
\newcommand{\myargmax}[1]{\mathop{\textrm{argmax}}_{#1}} 
\newcommand{\myst}{\textrm{s.t. }}

\newcommand{\denselist}{\itemsep -1pt}
\newcommand{\sparselist}{\itemsep 1pt}

\definecolor{pink}{rgb}{0.9,0.5,0.5}
\definecolor{purple}{rgb}{0.5, 0.4, 0.8}   
\definecolor{gray}{rgb}{0.3, 0.3, 0.3}
\definecolor{mygreen}{rgb}{0.2, 0.6, 0.2}

\newcommand{\cyan}[1]{\textcolor{cyan}{#1}}
\newcommand{\red}[1]{\textcolor{red}{#1}}  
\newcommand{\blue}[1]{\textcolor{blue}{#1}}
\newcommand{\magenta}[1]{\textcolor{magenta}{#1}}
\newcommand{\pink}[1]{\textcolor{pink}{#1}}
\newcommand{\green}[1]{\textcolor{green}{#1}} 
\newcommand{\gray}[1]{\textcolor{gray}{#1}}    
\newcommand{\mygreen}[1]{\textcolor{mygreen}{#1}}    
\newcommand{\purple}[1]{\textcolor{purple}{#1}}       

\definecolor{greena}{rgb}{0.4, 0.5, 0.1}
\newcommand{\greena}[1]{\textcolor{greena}{#1}}

\definecolor{bluea}{rgb}{0, 0.4, 0.6}
\newcommand{\bluea}[1]{\textcolor{bluea}{#1}}
\definecolor{reda}{rgb}{0.6, 0.2, 0.1}
\newcommand{\reda}[1]{\textcolor{reda}{#1}}

\def\changemargin#1#2{\list{}{\rightmargin#2\leftmargin#1}\item[]}
\let\endchangemargin=\endlist
                                               
\newcommand{\cm}[1]{}

\newcommand{\mhoai}[1]{{\color{magenta}\textbf{[MH: #1]}}}

\newcommand{\mtodo}[1]{{\color{red}$\blacksquare$\textbf{[TODO: #1]}}}
\newcommand{\myheading}[1]{\vspace{0.5ex}\noindent \textbf{#1}}
\newcommand{\htimesw}[2]{\mbox{$#1$$\times$$#2$}}

% The following are useful for creating homework or exams

%\newif\ifshowsolution
%%\showsolutionfalse
%%\showsolutiontrue
%
%\ifshowsolution  
%\newcommand{\Comment}[1]{\paragraph{\bf $\bigstar $ COMMENT:} {\sf #1} \bigskip}
%\newcommand{\Solution}[2]{\paragraph{\bf $\bigstar $ SOLUTION:} {\sf #2} }
%\newcommand{\Mistake}[2]{\paragraph{\bf $\blacksquare$ COMMON MISTAKE #1:} {\sf #2} \bigskip}
%\else
%\newcommand{\Solution}[2]{\vspace{#1}}
%\fi
%
%\newcommand{\truefalse}{
%\begin{enumerate}
%	\item True
%	\item False
%\end{enumerate}
%}
%
%\newcommand{\yesno}{
%\begin{enumerate}
%	\item Yes
%	\item No
%\end{enumerate}
%}

\newcommand{\Sref}[1]{Sec.~\ref{#1}}
\newcommand{\Eref}[1]{Eq.~(\ref{#1})}
\newcommand{\Fref}[1]{Fig.~\ref{#1}}
\newcommand{\Tref}[1]{Table~\ref{#1}}

\begin{abstract}
  We introduce a novel learning problem: decoding gaze into natural language descriptions of human goals across diverse visual tasks. Unlike prior work, which frames gaze decoding as a discriminative task over predefined categories, we formulate it as a generative learning problem: training a model to produce free-form descriptions that capture the rich nuances and open-ended nature of human intentions beyond fixed labels. To this end, we introduce \emph{Gazette}, the first gaze-to-text decoding framework. Based on multimodal large language models (MLLMs), Gazette learns to decode gaze scanpaths into natural language for goals that may extend beyond categorical labels and require articulation in natural language. To help Gazette filter out individual differences in gaze behavior and learn the goal-specific spatiotemporal dynamics crucial for generating accurate natural language goal descriptions, we propose a novel strategy that leverages the encyclopedic knowledge and reasoning abilities of a large language model to synthesize natural language explanations of goal-directed attentional behavior called \emph{think-aloud transcripts}. Instruction tuning on these synthetic narratives allows Gazette to achieve state-of-the-art performance in gaze decoding across multiple tasks, demonstrating its generalizability and versatility, thereby enabling gaze to serve as a powerful, non-intrusive cue for inferring human goals and intentions in diverse scenarios. Code is available at \url{https://github.com/cvlab-stonybrook/Gazette}.
\end{abstract}

\section{Introduction}
\label{sec:introduction}

%Gaze decoding offers a non-intrusive and practical means of inferring human attention and intent, %serving as a valuable alternative to neural decoding from EEG~\cite{hollenstein2021decoding, daly2023neural}, fMRI~\cite{xia2024dream, takagi2023high}, MEG~\cite{defossez2023decoding, benchetrit2023brain}, and ECoG~\cite{chestek2013hand, komeiji2024feasibility} that often need bulky, expensive, and invasive equipment. As eye-tracking technology becomes increasingly accessible and accurate, across both wearable and non-wearable devices, the utility of gaze decoding grows, enabling a wide range of applications including user engagement analysis~\cite{buhler2024task, khokhar2019eye}, driver monitoring~\cite{tawari2014driver}, human-robot interaction~\cite{li2017implicit}, hands-free assistive technologies~\cite{Perfect2020}, and psychological diagnosis~\cite{liaqat2021predicting}.
The ability to infer and analyze human attention and intention underpins a wide range of applications, including user engagement analysis~\cite{buhler2024task, khokhar2019eye}, driver monitoring~\cite{tawari2014driver}, human–robot interaction~\cite{li2017implicit}, hands-free assistive technologies~\cite{Perfect2020}, and psychological diagnosis~\cite{liaqat2021predicting}. One practical approach is to track a user's gaze and decode it to estimate their underlying intent and goals. This process, known as gaze decoding, offers a non-intrusive and scalable alternative to neural decoding methods based on EEG~\cite{hollenstein2021decoding, daly2023neural}, fMRI~\cite{xia2024dream, takagi2023high}, MEG~\cite{defossez2023decoding, benchetrit2023brain}, and ECoG~\cite{chestek2013hand, komeiji2024feasibility}, which typically require bulky, costly, and sometimes invasive equipment.

Gaze decoding is an important research problem, and existing approaches~\cite{barz2020visual, sattar2015prediction, sattar2017predicting, sattar2020deep, nishiyasu2024gaze, wang2024gazegnn, Mondal_2025_ICCV} formulate gaze decoding as a classification task over predefined categories, addressing questions such as ``Does the gaze behavior reflect free viewing or goal-directed behavior?'' (i.e., binary classification) or ``Which object is the user searching for?'' (i.e., multi-class classification over a fixed object set), with the model outputting a single label from a predefined category space.

However, such formulations are inherently constrained by a fixed label set. When the number of categories is small, outputs tend to be overly coarse and lack the fine-grained detail required for many downstream applications; when the label space is large, the model demands extensive annotated data to achieve adequate coverage. More fundamentally, any predefined category set is limited in expressive power and cannot capture the richness of natural language. In this work, we move beyond category-based gaze decoding and study the task of inferring a person's goal from attention sequences, where the output may range from simple open-vocabulary object labels (e.g., ``bowl'', ``car'') to free-form, context-rich descriptions such as ``red bowl to the left of the cup in the middle.''

This natural language-based formulation of gaze decoding enables a broad range of real-world applications. In e-commerce, descriptive language-based representations of gaze patterns allow systems to capture nuanced user intent when interacting with novel products that cannot be described by a fixed taxonomy. Instead of merely identifying attention to a ``chair,'' decoding gaze into descriptions such as ``mid-century walnut armchair with woven rattan back'' provides richer semantic signals about style, material, and aesthetic preference. In educational settings, gaze descriptions such as ``the equation's denominator term'' or ``the shaded region under the curve'' may reveal conceptual focus or misunderstanding more precisely than coarse object tags. 
In assistive technologies using augmented reality glasses, generating descriptions such as ``the user is approaching and looking at the top of the escalator'' can help infer navigation intent. By decoding gaze behavior into expressive natural language, the model unlocks fine-grained, open-vocabulary semantic signals that enable reasoning over new, rare, or previously unseen concepts and improve adaptability across domains.

To operationalize this formulation, we propose \emph{\textbf{Gaze}-\textbf{t}o-\textbf{te}xt} (\textit{Gazette}), a multimodal large language model (MLLM)-based generative framework that decodes goal-directed human attention across diverse gaze behaviors. Gazette takes as input an image $I$ and a language instruction ${T}_{GazeDec}$ textually representing the gaze scanpath $S$, and the output is text $D$ describing the \emph{cognitive context} of the human. We define cognitive context at two levels: a coarse \emph{behavior type} (visual search, object referral, or visual question answering) and a fine-level \emph{stimulus}, that triggers attentional movements. The latter ranges from a search target's category label to free-form text such as a referring expression or question.

Gazette is essentially an MLLM for the task of gaze decoding, yet the standard training paradigm for such models does not directly transfer. The conventional approach—instruction tuning~\cite{liu2023visual, li2024llara}—starts from an MLLM, constructs labeled instruction data, and fine-tunes the model using these supervision signals. In the context of gaze decoding, the instruction-tuning data would consist of $N$ triplets of an image, a gaze scanpath, and the corresponding description of the goal, $\{(I_i, S_i, D_i)\}_{i=1}^{N}$, collected by first defining a goal $D_i$ (e.g., ``red car on right'') for a given image $I_i$ and then recording the gaze behavior $S_i$ of a participant pursuing that goal. However, adapting an MLLM to perform gaze decoding—i.e., learning a mapping from an input image and scanpath $(I_i, S_i)$ to the goal description $D_i$—is fundamentally challenging because the supervision is inherently weak. Although the goal description $D_i$ reflects the task-driven objective, the observed gaze signal $S_i$ is not purely goal-driven; it is also substantially influenced by individual differences among participants. In other words, for the same goal $D_i$, different participants may exhibit different scanpaths $S_i$'s. The model must therefore learn a mapping from partially confounded inputs to clean task-level outputs. This disentanglement problem is further exacerbated by the lack of domain-specific priors in MLLMs~\cite{hamza2025llava, duan2024cityllava, mohbat2024llava}, particularly regarding gaze behavior, making it difficult to separate goal-relevant signals from participant-specific variation. As a result, models trained solely through standard instruction tuning exhibit suboptimal performance on the gaze decoding task.

Building on this insight, we argue that effective gaze decoding requires factoring out the goal-driven signal from participant-specific confounding factors. At first glance, this appears intractable—akin to solving an underdetermined system with a single equation and multiple unknowns. However, when multiple gaze behaviors are observed for the same visual stimulus under a shared goal, the goal-driven component remains invariant while participant-specific variation changes across instances. This shared invariance enables the common attentional objective to be isolated despite individual differences.

% Concretely, given scanpaths from multiple participants under a shared goal, we first generate \emph{think-aloud transcripts}\cite{ericsson1984protocol, vansomeren1994think}—narratives that highlight attentional patterns and information relevant to the top-down attentional goal while filtering out individual variability. These transcripts are generated by leveraging the reasoning capabilities of GPT-4\cite{achiam2023gpt}, which interprets the shared structure across scanpaths to extract the common \emph{top-down attention allocation strategy.} 
% The resulting transcripts are then used as auxiliary instruction-tuning data to train Gazette. Notably, multiple scanpaths and transcript generation are required only during training; at inference time, the trained model can decode the goal from a single scanpath. Empirically, incorporating the auxiliary think-aloud objective leads to substantial improvements in gaze decoding performance.

Concretely, given scanpaths from multiple participants sharing a goal, we generate \emph{think-aloud transcripts}\cite{ericsson1984protocol, vansomeren1994think}—narratives that surface goal-relevant attentional patterns while filtering out individual variability. We obtain these by prompting GPT-4\cite{achiam2023gpt} to interpret the shared structure across scanpaths and extract the common \emph{top-down attention allocation strategy}, then use the transcripts as auxiliary instruction-tuning data for Gazette. Crucially, multiple scanpaths and transcript generation are needed only during training; at inference, the model decodes the goal from a single scanpath. This auxiliary think-aloud objective yields substantial gains in gaze decoding performance.

In summary, the contributions of this paper are threefold. First, we introduce the task of unconstrained decoding of top-down attention, where goals are expressed in natural language. Second, we propose \textit{Gazette}, a text-generative MLLM-based framework designed for this task. Third, we introduce an auxiliary \emph{think-aloud transcript} prediction objective that encourages Gazette to capture goal-specific attentional patterns, thereby improving decoding performance across a broad range of top-down attentional tasks.

% In summary, our contributions are: 

% \begin{enumerate}
%     \item We introduce the task of unconstrained decoding of top-down attention, where goals are expressed in natural language, supporting a wide range of human gaze behaviors.%, including those with complex stimulus such as object referral and visual question answering.

%     \item We propose \textit{Gazette}, a novel text-generative MLLM-based framework, which is instruction tuned to decode a scanpath of gaze fixations to natural language. 

%     \item We further enhance Gazette by instruction tuning on an auxiliary \emph{think-aloud transcript} generation task for identifying goal-specific attentional patterns within scanpaths. 

%     \item We derive ground truth for think-aloud transcripts by prompting GPT-4 using a novel prompting strategy that exploits commonalities in gaze behavior of multiple observers engaged in the same top-down attentional task and visual stimulus. 
%     \end{enumerate}
\section{Related Work}
\label{sec:related_work}

\textbf{Goal-directed Human Attention. } Goal-directed attention, in contrast to bottom-up attention~\cite{IttiPAMI98, masciocchi2009everyone}, is the top-down control exerted by frontal-parietal brain areas that modulates processing of sensory input based on current task demand, prior knowledge and expectations~\cite{henderson2007visual, koehler2014saliency}. %Most research on human gaze behavior has been on bottom-up attention or free-viewing~\cite{IttiPAMI98, masciocchi2009everyone, berg2009free}---a rapid, stimulus-driven process that highlights salient sensory inputs via early visual cortices. On the other hand, top-down or goal-directed attention~\cite{henderson2007visual, koehler2014saliency} is a slower mechanism %originating from frontal–parietal networks that modulates sensory processing based on prior knowledge, task demands, and expectations. %In recent years, 
Several datasets have been collected to study various facets of goal-directed gaze behavior. %Chen \etal~\cite{chen2021coco} collected 
COCO-Search18~\cite{chen2021coco} is a dataset of visual search gaze fixations from 10 participants performing Target-Present/Absent tasks on 6,202 natural images spanning 18 target categories. %Relatedly, Mondal \etal~\cite{mondal2025look} collected 
RefCOCO-Gaze~\cite{mondal2025look} contains gaze scanpaths from 220 participants viewing 2,094 COCO images while simultaneously hearing referring expressions grounding objects in the images. AiR-D~\cite{chen2020air} is a dataset with scanpaths of 20 participants performing the visual question answering task for 195 image–question pairs. In this study, we use COCO-Search18, RefCOCO-Gaze and AiR-D.

\textbf{Gaze Decoding.} Gaze measured by eye-trackers can be a noninvasive means of understanding human intention. %In his classical work, 
Yarbus~\cite{yarbus1967eye} showed that eye movements depend strongly on the viewer's task, with fixations varying across instructions for the same visual stimulus. Building on this, prior work has decoded goal-directed tasks from eye movements~\cite{zelinsky2008eye, zelinskyEyeCan2013, bahle2017human, borji2014defending, borji2015eyes}, reconstructed images from fixations~\cite{wang2019mental, strohm2021neural, strohm2023usable, strohm2023facial}, and inferred user tasks or activities from gaze~\cite{bulling2010eye, bektacs2024gaze, hu2021ehtask, steil2015discovery, chen2022characterizing}. For search-target inference specifically, Sattar et al.~\cite{sattar2015prediction, sattar2017predicting, sattar2020deep} addressed both images and categories, and later methods used pre-trained CNN features~\cite{stauden2018visual, barz2020visual}.
More recent models add structure: GST~\cite{nishiyasu2024gaze} fuses category semantics with scanpaths to predict COCO-Search18 search targets, and GazeGNN~\cite{wang2024gazegnn}, a radiological diagnosis model, can be adapted for gaze target decoding. All frame decoding as classification -- selecting one goal from a fixed set -- limiting real-world applicability. %These related gaze decoding models either  decode mental images, or user intentions or goals in the form of distinct categories that severely constrain their applicability to real-life applications. 
Relatedly, Chen \etal~\cite{chen2024gazexplain} jointly predicted scanpaths and per-fixation natural-language explanations, but took scanpaths as output rather than input. Our work instead accepts human scanpaths as input and decodes the cognitive context behind top-down attention. Mondal \etal~\cite{Mondal_2025_ICCV} enabled zero-shot search-target decoding via gaze–language alignment, but were limited to a fixed vocabulary of target descriptions at inference. More recently, Xue et al.~\cite{xue2026personalized} absorbed training-time attention into a subject embedding to personalize image descriptions, conditioning on \emph{who} is describing rather than decoding a goal, and thus requiring no gaze at inference. Gazette instead takes the scanpath as input and decodes the underlying goal, treating individual variation as a confound to filter out. To our knowledge, Gazette, is the first generative, gaze-to-text decoding framework.  

\begin{figure*}
  \centering
  \includegraphics[width=\linewidth]{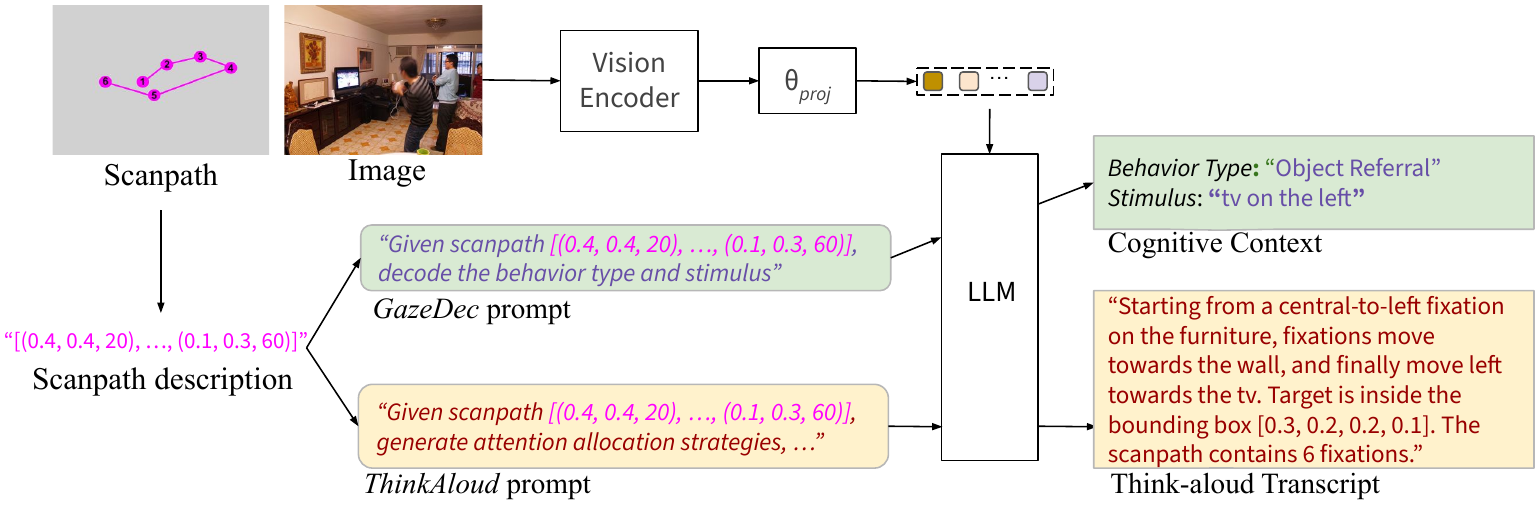}
  \caption{\textbf{\emph{Gazette}}: A text-generative decoding framework for top-down attention. For an input of an image and a language instruction conveying scanpath information, a textual response is generated by the Multimodal LLM comprising a Vision Encoder, an LLM, and a linear projection $\theta_{proj}$ interfacing the Vision Encoder and the LLM. Language instruction can correspond to either the primary gaze decoding task \emph{GazeDec}, or the auxiliary think-aloud transcript generation task \emph{ThinkAloud}.}
  \label{fig:gazette}
\end{figure*}
\textbf{Instruction Tuning of MLLMs. }Instruction tuning or supervised fine-tuning (SFT) for multimodal large language models (MLLMs) adapts pre-trained vision-language architectures for downstream tasks by training on instruction-response pairs, enhancing model performance on specific tasks and general instruction-following~\cite{liu2023visual,liu2024improved, ranasinghe2024learning}. %Recent m
Methods like LLaVA \cite{liu2023visual} and VIGC \cite{wang2024vigc} generate large-scale visual instruction datasets, often with proprietary LLMs like GPT-4 endowed with image context expressed via object bounding box annotations in the scene. Visual instruction tuning has been applied in diverse domains. Instruction-tuned MLLMs have excelled in robotics~\cite{driess2023palm, li2024llara}, healthcare~\cite{singhal2023clinical}, and e-commerce~\cite{liu2024improved}, demonstrating how instruction tuning transforms general-purpose MLLMs into specialized, domain-adapted agents. Inspired by these works, we curate the first instruction tuning dataset for gaze decoding, enabling the adaptation of general-purpose MLLMs for the decoding of gaze behavior.

\section{\emph{Gazette}: Gaze-to-Text Decoding of Human Attention}
\label{sec:method}

% Existing gaze decoding methods for visual search tasks either adopt an $K$-way classification framework to select one of the $K$ predefined search target categories~\cite{barz2020visual, nishiyasu2024gaze}, or learn gaze-target embedding alignment, but have to be provided a user-defined vocabulary of possible search targets to choose from during inference~\cite{Mondal_2025_ICCV}.  However, this severely restricts the applicability and extensibility of these methods to a wider range of gaze behaviors, particularly those that have complex attribute and contextual information about objects, \eg object referral~\cite{mondal2025look} and VQA~\cite{chen2020air}, %therefore necessitating 
% needing precise natural language specification. If one were to adapt Mondal \etal~\cite{Mondal_2025_ICCV}'s method for object referral or VQA, they will have to define candidate referring expressions/questions during inference -- a requirement that is both highly non-trivial and impractical for natural scenes. 
% \mhoai{The above paragraph seems redundant, repeating the intro? It is also make this overview paragraph of this method section overly long.}

\textit{Gazette} is an MLLM-based generative framework that decodes diverse gaze behaviors into natural language. Specifically, Gazette extends the LLaVA-1.5-7B foundation model~\cite{liu2024improved} and is fine-tuned on our novel visual instruction-tuning dataset, which consists of tasks closely aligned with gaze understanding and is described in detail in this section. The input to Gazette consists of an image and a gaze scanpath, and the output is a hierarchical representation capturing two facets of top-down attentional control: (i) a coarse-level \emph{behavior type}, representing the category of goal-directed behavior (e.g., Target-Present Visual Search, Target-Absent Visual Search, object referral, or VQA), and (ii) a fine-level \emph{stimulus}, specifying the concrete top-down goal. For visual search, the stimulus corresponds to the target category (e.g., ``car''); for object referral, it corresponds to a referring expression (e.g., ``red car on the right''); and for VQA, it corresponds to the question itself (e.g., ``Is there a red car on the road?'').

The remainder of this section is organized as follows. We first describe the architecture of Gazette and its processing pipeline. We then present \emph{think-aloud transcript generation}, our proposed method for extracting narratives that highlight goal-relevant attentional patterns while filtering out individual variability. Finally, we discuss the training and inference procedures.

\subsection{Architecture and Decoding Pipeline}

Built upon the LLaVA foundation model~\cite{liu2024improved}, Gazette adopts its multimodal architecture, consisting of a \emph{Vision Encoder}\cite{radford2021learning} coupled with a large language model (LLM). The input image $I \in \mathbb{R}^{3\times H\times W}$ is processed by the Vision Encoder, and the resulting visual features are projected via a lightweight MLP $\theta_{proj}$ into the LLM’s token embedding space. A gaze scanpath $S = \{f_i \mid i=0,\ldots,N-1\}$ is a sequence of $N$ fixations $f_i = (x_i, y_i, t_i)$, where $(x_i, y_i)$ denotes the normalized spatial coordinates (scaled to $[0,1]$) and $t_i$ the fixation duration. Following prior visual instruction-tuning work\cite{li2024llara, liu2024improved, ranasinghe2024learning}, we encode $S$ textually using a prompt template to produce a language instruction $T_{GazeDec}$ (e.g., ``Given scanpath $[(0.4, 0.4, 20), \ldots, (0.1, 0.3, 60)]$, decode the behavior type and stimulus''). The projected visual tokens and textual prompt are concatenated and fed into the LLM, which autoregressively generates natural language output $D$ describing the cognitive context. We refer to this primary decoding task as \emph{GazeDec}. The generated response $D$ can be parsed into two components: $D_{type}$, representing the behavior type, and $D_{goal}$, specifying the stimulus or goal. The components of Gazette and the overall processing pipeline are illustrated in~\Fref{fig:gazette}.

In addition to the primary decoding task \emph{GazeDec}, Gazette also supports an auxiliary think-aloud task that decodes the attention allocation strategy (please find more detailed description in the next subsection). For this task, the inputs remain the image $I$ and the scanpath $S$, but the prompt template is modified to produce $T_{ThinkAloud}$, which textualizes the scanpath and instructs the model to generate an attention allocation strategy (e.g., ``Given scanpath $[(0.4, 0.4, 20), \ldots, (0.1, 0.3, 60)]$, generate the attention allocation strategy \ldots''). The model then autoregressively generates the corresponding think-aloud transcript $D_{TaT}$. This auxiliary task is also illustrated in~\Fref{fig:gazette}.

\subsection{ThinkAloud Transcript Generation}
\label{sec:genattstrat}

Assume we are given training data consisting of $N$ triplets of an image, a gaze scanpath, and the corresponding goal description, ${(I_i, S_i, D_i)}_{i=1}^{N}$. The primary gaze decoding task is to learn a mapping from $(I_i, S_i)$ to $D_i$. However, defining this mapping as an MLLM and fine-tuning it solely on the primary objective yields suboptimal performance, because the observed scanpath is not purely goal-driven, and general-purpose MLLMs lack prior knowledge of human attentional control, making it difficult to disentangle goal-specific signals from individual variability in scanpaths. This difficulty can be viewed through the lens of an underdetermined inverse problem. For a single participant, the observed scanpath $S_i$ can be conceptualized as a function of both the latent goal $D_i$ and participant-specific factors $U_i$, i.e., $S_i = f(I_i, D_i) + U_i$. Although supervision is provided for $D_i$, the nuisance variable $U_i$ remains unobserved and entangled with the goal signal, making it difficult to recover a clean mapping from $(I_i, S_i)$ to $D_i$. In this setting, attempting to infer $f(I_i, D_i)$ from a single equation involving both $f(I_i, D_i)$ and $U_i$ is fundamentally ill-posed. However, when multiple scanpaths $S_i$'s are observed for the same visual stimulus under a shared goal, the participant-specific factors vary while the underlying goal remains invariant. These multiple observations introduce additional structure, allowing the goal-related signal to be isolated from individual variation. Intuitively, aggregating diverse yet goal-consistent behaviors makes the disentanglement of the shared objective feasible.

Suppose the training data can be partitioned into groups such that if indices~$i$ and $j$ belong to the same group $G$, then $I_i = I_j$ and $D_i = D_j$, while $S_i \neq S_j$ because the scanpaths are collected from different participants. With slight abuse of notation, we denote such a group as $(I, D, \{S_i \mid i \in G\})$, where $I$ and $D$ represent the shared stimulus and goal. Although individual scanpaths vary, we posit that they share a common component attributable to the goal-driven attention process, which we interpret as the underlying \emph{attention allocation strategy}. This hypothesis is grounded in classic findings that eye movement patterns vary systematically with task demands~\cite{yarbus1967eye}, yet are simultaneously influenced by multiple factors, including bottom-up salience and idiosyncratic viewer characteristics. Beyond this descriptive observation, the convergence we exploit has a principled theoretical basis. Najemnik \& Geisler's Bayesian ideal-observer model~\cite{najemnik2005optimal} shows that the optimal fixation policy depends on the target template, the scene prior, and the observer's visibility map; the first two are shared across observers performing the same task, while only the third varies mildly between individuals. Cross-observer convergence on a goal-carrying fixation pattern is therefore a theoretical prediction rather than a mere empirical regularity, with individual idiosyncrasies acting as approximately orthogonal noise. Hayhoe \& Ballard's Cognitive Relevance Theory~\cite{hayhoe2005eye} extends this account beyond visual search through a shared, goal-determined relevance signal, predicting the same convergence for object referral and VQA. Thus, while isolating goal-specific information from a single scanpath is difficult, the shared structure across multiple participants engaged in the same task provides a principled basis for extracting the common goal-driven signal.

To operationalize this idea, we leverage GPT-4~\cite{achiam2023gpt} to derive \emph{think-aloud transcripts} that capture this shared attention allocation strategy. Each transcript is structured into three idea units—(i) a top-down attention allocation explanation, (ii) the inferred target location, and (iii) the scanpath length—following principles from think-aloud protocol analysis~\cite{fonteyn1993description}. These transcripts serve as supervision for the auxiliary \emph{ThinkAloud} task.

\def\subFigSz{0.24\linewidth}
\begin{figure*}[t]
  \centering
  \includegraphics[width=\linewidth]{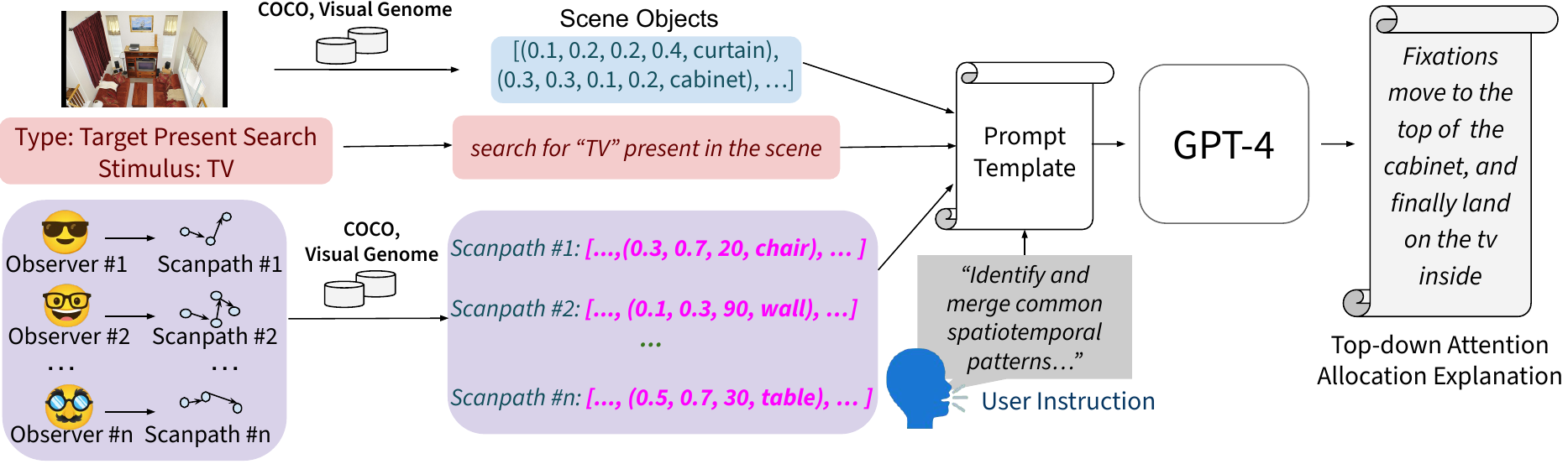}
            \makebox[\subFigSz][h]{(a)}\\
  \makebox[\subFigSz][h]{}\\
  \makebox[\subFigSz][h]{ \includegraphics[width=\subFigSz]{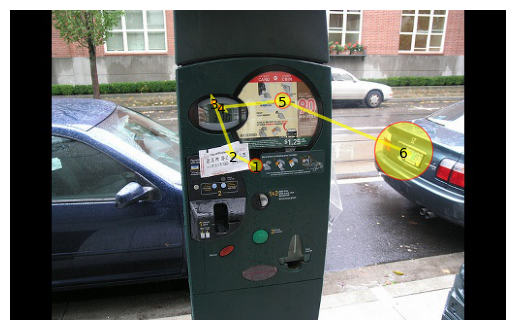}}
 \makebox[\subFigSz][h]{ \includegraphics[width=\subFigSz]{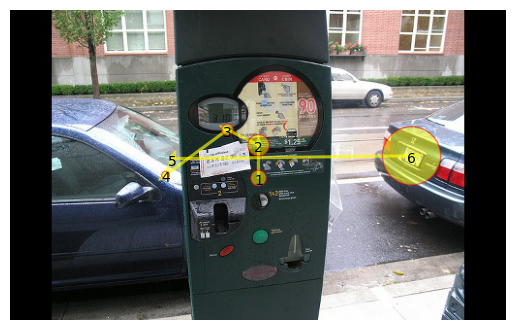}}
 \makebox[\subFigSz][h]{ \includegraphics[width=\subFigSz]{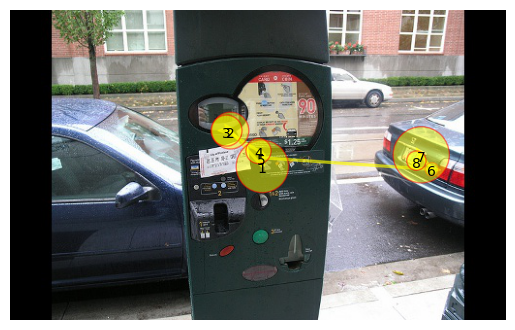}}
 \makebox[\subFigSz][h]{ \includegraphics[width=\subFigSz]{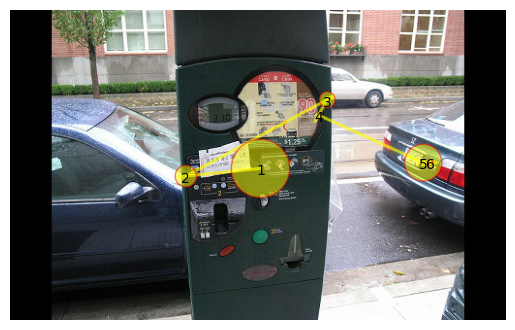}}\\
 \makebox[\subFigSz][h]{\small{For stimulus: \textit{Object Referral} for \textit{``car on the right''}.}}\\
\begin{minipage}[t]{\linewidth}{\small{GPT-4 generated: {\it Most humans initially fixate on the large vertical parking meter spanning a significant portion of the left-central part of the image, then shift focus toward the large car located on the right side, spending varying durations there.}}}
\end{minipage}\\
(b)
  \caption{A novel prompting strategy for GPT-4 to extract top-down attention allocation explanation for think-aloud transcript generation task (\emph{ThinkAloud}). (a) This strategy instructs GPT-4 to summarize common spatiotemporal patterns in scanpaths of $n$ participants, given the task type and goal, along with scene information via scene object bounding boxes from COCO~\cite{linMicrosoftCoco2014} and Visual Genome~\cite{krishna2017visual}. The response is used to construct the think-aloud transcript. (b) Sample scanpaths from several users for an image-stimulus pair and the
corresponding GPT-4 generated response. }
  \label{fig:gpt4_extract}
\end{figure*}

%\myheading{Novel prompting strategy for GPT-4 to annotate top-down attention allocation explanation.} 
%In addition to the attentional goal, human top-down attention control is affected by several individual factors such as age, ocular health, and neurodiversity. %Drawing analogy to paintings and text, These individual factors are analogous to ``style'' while the goal is analogous to ``content''.  We hypothesize that this goal-specific information is contained within the \emph{common spatiotemporal patterns} within scanpaths of multiple participants engaged in the same attentional task for the same visual stimulus, \eg image. These common patterns are likely due to the common goal shared by the participants, potentially unaffected by individual-level idiosyncrasies and noise. % (if any, due to reasons like eye-tracking inaccuracies).    
We employ GPT-4~\cite{achiam2023gpt} to analyze scanpaths from multiple participants who share the same cognitive context for a given image and to summarize their common spatiotemporal patterns (see Fig.~\ref{fig:gpt4_extract}). We select GPT-4 for its strong capability in identifying latent patterns in structured data\cite{mirchandani2023large,weber2024pattern} and its extensive world knowledge. As noted in existing literature~\cite{wang2024vigc}, currently available multimodal large language models (MLLMs) remain less capable and more costly than their text-only LLM counterparts. Therefore, following LLaVA~\cite{liu2023visual}, we adopt text-only GPT-4 for its accessibility, cost-effectiveness, and computational efficiency. We provide scene context in the form of bounding-box annotations of objects in the scene derived from COCO~\cite{linMicrosoftCoco2014} for COCO-Search18, RefCOCO-Gaze, and COCO-originated images in the AiR-D dataset, and from Visual Genome~\cite{krishna2017visual} scene graphs for the Flickr-originated images in the AiR-D dataset sourced from the GQA dataset~\cite{hudson2019gqa}. Again, to counter scale variation, we normalize bounding-box parameters. Additionally, we provide scanpath information for all observers in the form of normalized fixation co-ordinates, raw fixation durations, and category labels of the fixated objects (derived from COCO and Visual Genome bounding box annotations). Finally, cognitive context is also included in the prompt. %Intricate details about the prompt can be found
Prompt details are given in the supplement. 

GPT-4's response reflects common patterns across scanpaths of several participants, which we hypothesize indicates the top-down attentional allocation strategies utilized by the human visual system, therefore serving as pseudo-annotation for the ``top-down attention allocation explanation'' idea unit of the think-aloud transcript. Because this response captures the \emph{common} patterns, it remains consistent across \emph{all} scanpaths from multiple participants, %therefore serving 
and can be used as pseudo-annotation for scanpaths from \emph{each} individual participant.
Additionally, following previous visual instruction tuning research~\cite{ranasinghe2024learning, li2024llara} adapting MLLMs for specialized tasks, we propose two fundamental scanpath comprehension tasks as ``idea units'' in the think-aloud transcript: target localization, and counting fixations in a scanpath. Since MLLMs struggle in object localization~\cite{ranasinghe2024learning}, a crucial task underlying gaze behavior for visual search and object referral~\cite{yang2022target, mondal2023gazeformer}, we posit that Gazette will benefit from learning target localization, particularly for Visual Search and Object Referral tasks. Following previous MLLM literature~\cite{ranasinghe2024learning, liu2024improved, li2024llara}, we normalize bounding box parameters $[x,y,w,h]$ (where $x,y$ locate the upper-left corner, and $w$ and $h$ are the width and height of the bounding box, respectively) to deal with scale variation. This task also accounts for target-absent cases (null scenario), where Gazette must predict the absence of the target instead of a bounding box. The second task is inspired by previous research~\cite{tsang2010eseetrack, williams2019changing,sabab2022vis} suggesting that scanpath length indicates many properties of gaze behavior, \eg the degree of exploratory behavior compared to focused behavior. Prompts for both \emph{GazeDec} and \emph{ThinkAloud} along with human evaluation of the GPT-4-generated top-down attention allocation explanation pseudo-annotations are in the supplement.

\subsection{Model Training and Inference}

We detail Gazette's training and inference procedures below, with more details in the supplement.

\myheading{Training.} We initialize our model with pre-trained LLaVA-1.5-7B model~\cite{liu2024improved} weights, and fine-tune it via Low-Rank Adaptation (LoRA)~\cite{hu2021lora}, reducing the number of trainable parameters, mitigating overfitting on limited gaze training data. We avoid newer MLLMs to ensure fair comparison and isolate the impact of think-aloud transcript–based tuning from backbone improvements. %, which freezes the base model's weights and adds trainable, low-rank update matrices to every layer. 
 %This reduces the number of trainable parameters, mitigating overfitting on limited gaze training data. %Following maximum likelihood training for causal language models, w
We use an auto-regressive language modeling 
objective~\cite{liu2023visual} for instruction tuning on both \emph{GazeDec} and \emph{ThinkAloud} tasks. 

\myheading{Inference.} We prompt Gazette using $T_{GazeDec}$ and decode the response text using a greedy decoding strategy. The response text is then parsed to obtain the gaze behavior type $D_{type}$ and the stimulus $D_{goal}$. $D_{type}$ is encoded using a language encoder~\cite{wang2020minilm} and matched to label vocabulary by computing cosine similarities between each text embedding and each label's language embedding, where label vocabulary is \{\texttt{``Target-Present Search''}, \texttt{``Target-Absent Search''}, \texttt{``Object Referral''}, or \texttt{``Visual Question Answering''}\}. Similarly, for %visual search scanpaths from 
COCO-Search18, ${D}_{goal}$ is matched with the 18 target categories of the dataset. %from COCO-Search18. 
% The response templates for a scanpath with $N$ fixations are as follows: 

% \begin{enumerate}
%     \item \emph{BasicComp}\textsubscript{count}: \texttt{The scanpath contains [N] fixations.}
%     \item \emph{BasicComp}\textsubscript{loc}: We use different templates for our 4 tasks:
    
%     \begin{enumerate}
%         \item Target-Present Visual Search for target category name \texttt{[target]} in normalized bounding box \texttt{[bbox]} : \texttt{The human is searching for the <target> present in the image. The [target] is inside the bounding box [bbox]. }
%         \item Target-Absent Visual Search for target category name \texttt{[target]}: \texttt{The human is searching for the [target] absent in the image.}
%         \item Object Referral for target in normalized bounding box \texttt{[bbox]}: \texttt{The human is grounding the target described by a referring expression. The target is inside the bounding box [bbox].}
%         \item Visual Question Answering: \texttt{The human is answering a question about the image.}
%     \end{enumerate}
    
% \end{enumerate}

\section{Experiments}
\label{sec:experiments}

\emph{Gazette} can decode gaze scanpaths for multiple gaze behaviors, including categorical visual search, object referral, and visual question answering (VQA). In this section, we evaluate Gazette's gaze decoding capabilities using a wide range of metrics across multiple gaze behaviors. Gaze behavior type prediction is trivial for Gazette (see supplement), perhaps due to artifacts in scanpaths from distinct %datasets having 
data collection setups. Here, we focus on stimulus decoding.
%Our final method 
We train our model on two NVIDIA RTX A6000 GPUs with a total batch size of 32 and a learning rate of 2e-5. Training is done on the training splits of COCO-Search18 (both Target-Present and Target-Absent trials), RefCOCO-Gaze and AiR-D, and evaluation is done on the corresponding test splits. %We trained \emph{Gazette}  using a combination of primary task instructions for \emph{GazeDec} and auxiliary task instructions for \emph{ThinkAloud} task. 
To evaluate and analyze the efficacy of our proposed auxiliary task, we compare performances of our full model Gazette trained on both sets of instructions $\mathbf{T}_{GazeDec}$ and $\mathbf{T}_{ThinkAloud}$  (corresponding to \emph{GazeDec} and \emph{ThinkAloud}, respectively), and a variant where the MLLM is trained only on $\mathbf{T}_{GazeDec}$ but not $\mathbf{T}_{ThinkAloud}$. The latter variant is simply denoted as ``\emph{w/o ThinkAloud}''. %We probe the idea units of think-aloud transcripts in the supplementary material. 

While discriminative gaze decoding methods exist~\cite{wang2024gazegnn, nishiyasu2024gaze, Mondal_2025_ICCV}, Gazette is the first generative gaze decoding model to our knowledge, so we constructed baselines based on the LLaVA~\cite{liu2024improved} model. For Object Referral and Target-Present Visual Search, a majority of last fixations land on the target, so we fine-tune LLaVA-1.5 to describe the object where the final fixation lands, to yield the gaze stimulus. We call this baseline \emph{LLaVA-{last}}. %Note that we further disentangle 
%material. 
Another baseline is a frozen \emph{LLaVA-1.5}~\cite{liu2024improved} prompted with scanpath information and instructions to describe the goal.
We also compare Gazette with Mondal \etal~\cite{Mondal_2025_ICCV}, GST~\cite{nishiyasu2024gaze} and GazeGNN~\cite{wang2024gazegnn} (adapted by us for visual search gaze decoding) in visual search gaze decoding with COCO-Search18~\cite{chen2021coco}. More details of Gazette and the aforementioned baselines are in the supplement.

\subsection{Evaluation Metrics}

We curated metrics suited to each gaze behavior. Visual search is decoded as a target category from a predefined set, whereas object referral and VQA are decoded into free-form expressions and questions; the metrics differ accordingly.
\subsubsection{Text Generation evaluation metrics for Object Referral and VQA.} 

We evaluate model text-generative capabilities on Object Referral gaze decoding and VQA gaze decoding using two paradigms: (1) using standard lexical overlap metrics, and (2) using GPT-4 to assess generated texts using a set of abstract rubrics, also known as LLM-as-a-Judge paradigm. %We detail these two paradigms below.

    \textbf{\textit{Lexical Overlap Metrics}}:  To assess the text generation-based decoding capabilities of Gazette, we use a broad set of standard metrics used to evaluate the lexical overlap of generated text from text generation models with the ground truth text. \textbf{BLEU-1 to BLEU-4}~\cite{papineni-etal-2002-bleu} evaluate machine‑generated text by comparing its n‑gram precision (n=1--4) against reference translations and apply a brevity penalty to discourage overly short output. \textbf{METEOR}~\cite{banerjee-lavie-2005-meteor} aligns candidate and reference translations via exact, stem, synonym, and paraphrase matches, then computes a recall-weighted harmonic mean of unigram precision and recall with a fragmentation penalty to preserve word order. \textbf{ROUGE-L}~\cite{lin-2004-rouge} measures the longest common subsequence between candidate and reference texts, combining recall and precision into an F‑score to capture sentence‑level structure and word‑order overlap. \textbf{CIDEr}~\cite{vedantam-etal-2015-cider} captures human consensus by computing the TF–IDF–weighted cosine similarity of n‑gram vectors between a candidate and multiple references, highlighting terms frequent in the candidate and distinctive in the references. While there is only one reference question for VQA samples in AiR-D, there are multiple reference referring expressions in RefCOCO~\cite{yu2016modeling} annotated by multiple annotators for the same object in an image. This allows us to compute lexical overlap metrics for each referring expression as the candidate, while reserving the others as ground truth. This is repeated for all referring expressions; the average defines the RefCOCO Inter-Annotator Consistency (\textbf{RefCOCO-IAC}), a noise ceiling approximating human annotator variability (similar to Inter-Observer Consistency in behavioral literature).

\textbf{\textit{LLM-as-a-Judge}}: In this evaluation paradigm popularized by recent text-generative research~\cite{liu2023visual, zheng2023judging}, a very large language model (such as GPT-4) is used as an external evaluator to assess the quality of the generated outputs, providing a scalable and cost-effective alternative to human evaluation. Additionally, this allows more interpretable evaluation rubrics to be used (\eg fluency, helpfulness, relevance), allowing for a more nuanced, and context-aware, approach to assessing model-generated text that might not be possible with standard lexical overlap metrics. Note that in this evaluation setting, GPT-4 is used only to evaluate predicted referring expressions/questions against ground truth from RefCOCO-Gaze/AiR-D, not the think-aloud transcripts, avoiding any circular dependency via GPT-4 that may bias evaluation. 

For each scanpath, GPT-4 is prompted to evaluate texts using a scale from 1 to 10 for each rubric, where 1 is poor and 10 is excellent.  %applying the LLM-as-a-Judge paradigm to evaluating referral and VQA generative decoding capabilities for 
The prompt also contains these key information: \textbf{(1)} referring expressions/questions generated by all models to be evaluated, \textbf{(2)} ground truth referring expressions/questions, \textbf{(3)} scene context in the form of bounding boxes of every object in the scene (similar to the method described in Sec.~\ref{sec:genattstrat}), \textbf{(4)} rubrics to evaluate the generated texts on. The rubrics are distinct for object referral and VQA and are detailed below, and the detailed prompt to GPT-4 is provided in the supplement. 

For object referral, the rubrics are: (1) \textbf{Expression Overlap} - How much does the generated referring expression match the ground truth referring expressions, especially in terms of coverage of the entities, their attributes and spatial relationships? (2) \textbf{Referential Equivalence} - Does either of the ground truth expressions refer to the same object in the image as the generated expression? (3) \textbf{Category Correctness} - Is the type or category of the referred object mentioned correctly? 

On the other hand, for VQA, the rubrics are: (1) \textbf{Question Overlap} - How much does the generated question match the ground truth question, especially in terms of coverage of the entities, their attributes and spatial relationships?
(2) \textbf{Answer Equivalence} - Does the generated question and the ground truth question have the same correct answer on the image? 

\subsubsection{Target category prediction metrics for Visual Search Tasks.}

For COCO-Search18's 18 target categories, we assess decoding via \textbf{precision}, \textbf{recall}, \textbf{F$_1$ score}, and \textbf{accuracy}. As the test set is imbalanced across categories, we macro-average precision, recall, and F$_1$ over the 18 categories, weighting each equally. Accuracy ignores this imbalance, but we report it to compare with state-of-the-art methods Mondal \etal~\cite{Mondal_2025_ICCV} and GST~\cite{nishiyasu2024gaze}, neither of which has a public implementation. Mondal \etal report precision, recall, and F$_1$ only on the combined target-present/target-absent test set, not individually.

\subsection{Results}

We evaluate the efficacy of Gazette in decoding gaze scanpaths from four top-down attention tasks: Task A -- Object Referral, Task B -- Visual Question Answering (VQA), Task C -- Target-Present Visual Search, and Task D -- Target-Absent Visual Search. We evaluate generative gaze decoding on Object Referral (using RefCOCO-Gaze~\cite{mondal2025look} dataset) and VQA (using AiR-D~\cite{chen2020air} dataset) tasks under two paradigms: (1) Lexical Overlap Metric-based evaluation (Table~\ref{tab:lexical_res}) and (2) LLM-as-a-Judge-based evaluation (Table~\ref{tab:gpt4_eval}).

% \begin{enumerate}
%   \item \textbf{BLEU-1 to BLEU-4}~\cite{papineni-etal-2002-bleu}: BLEU (Bilingual Evaluation Understudy) metric evaluates machine‑generated text by comparing its n‑gram precision (n=1–4) against reference translations and applies a brevity penalty to discourage overly short outputs.
%   \item \textbf{METEOR}~\cite{banerjee-lavie-2005-meteor}: The METEOR (Metric for Evaluation of Translation with Explicit ORdering) metric evaluates machine-generated text by aligning candidate and reference translations using exact, stem, synonym, and paraphrase matches, then scores them via a recall‑weighted harmonic mean of unigram precision and recall. It applies a fragmentation penalty to penalize disordered matches, ensuring sentence‑level word‑order preservation.
%   \item \textbf{ROUGE-L}~\cite{lin-2004-rouge}: The Recall-Oriented Understudy for Gisting Evaluation – Longest Common Subsequence (ROUGE-L) metric measures the longest common subsequence between candidate and reference texts, combining recall and precision into an F‑score to capture sentence‑level structure and word‑order overlap.
%   \item \textbf{CIDEr}~\cite{vedantam-etal-2015-cider}: The CIDEr (Consensus-based Image Description Evaluation) metric captures human consensus in natural language by computing the TF–IDF–weighted cosine similarity of n‑gram vectors between a candidate caption and multiple reference captions, highlighting terms that are both frequent in the candidate and distinctive among the references.
% \end{enumerate}

%Note that f

\begin{table*}[ht!]
\centering
\caption{Performance of Gazette and baselines on Task A -- Object Referral Gaze Decoding evaluated using RefCOCO-Gaze~\cite{mondal2025look}, and Task B -- VQA Gaze Decoding evaluated using AiR-D~\cite{chen2020air} on the basis of lexical overlap metrics (B-1 through B-4 denote BLEU-1 through BLEU-4, M denotes METEOR, R-L denotes ROUGE-L, and C denotes CIDEr). 
%B-1 through B-4 denote BLEU-n metrics, M denotes METEOR, R-L denotes ROUGE-L, and C denotes CiDeR. 
Best results are highlighted in bold. Results exceeding RefCOCO Inter-Annotator Consistency (RefCOCO-IAC) values are underlined. Percentage improvements of Gazette over variant w/o \textit{ThinkAloud} are provided in parentheses.}
\label{tab:lexical_res}
\setlength{\tabcolsep}{0.1pt}
\begin{tabular}{llccccccccc}
% \hline

\toprule 
% &  & \multicolumn{4}{c}{\textbf{\small BLEU-n}}   &  \multirow{2}{*}{\scriptsize METEOR} & \\
%  \cmidrule{3-6}
\textbf{Task} & \textbf{Method} & {\bf \scriptsize B-1} & {\bf \scriptsize B-2} & {\bf \scriptsize B-3} & {\bf \scriptsize B-4} & {\bf \scriptsize M} & \textbf{\scriptsize R-L} & \textbf{\scriptsize C} \\
%\hline\addlinespace[2pt]
\midrule 
\multirow{6}{1cm}{Object Referral (Task A) } & RefCOCO-IAC & 0.500 & 0.282 & 0.148 & 0.057 & 0.244 & 0.452 & 1.115\\
\cmidrule{2-9}
& LLaVA-1.5~\cite{liu2024improved} & 0.066 & 0.018 & 0.006 & 0.0 & 0.077 & 0.105 & 0.062\\
 & \emph{LLaVA-{last}}& 0.170 & 0.080 & 0.039 & 0.014 & 0.113 & 0.221 & 0.113\\
\cmidrule{2-9}
 & \emph{w/o ThinkAloud} & 0.479 & 0.263 & 0.145 & \underline{0.070} & 0.232 & 0.443 & 0.872 \\
% & \emph{w/o GenAttStrat} & 0.493 & \underline{0.296} & \underline{0.153} & \underline{0.085} & 0.243 & \underline{0.458} & 0.937 \\
% &\emph{w/o BasicComp} & 0.472 & 0.259 & 0.134 & \underline{0.061} & 0.227 & 0.426 & 0.838 \\
&\emph{Gazette} & \underline{\textbf{0.519}} & \underline{\textbf{0.305}} & \underline{\textbf{0.175}} & \underline{\textbf{0.098}} & \underline{\textbf{0.248}} & \underline{\textbf{0.480}} & \textbf{0.974}\\
& & \scriptsize{(+8.36\%}) & \scriptsize{(+15.97\%}) & \scriptsize{(+20.69\%}) & \scriptsize{(+40.00\%}) & \scriptsize{(+6.90\%}) & \scriptsize{(+8.36\%}) & \scriptsize{(+11.70\%})

 \\\addlinespace[2pt]
\hline\addlinespace[2pt]

\multirow{4}{1.3cm}{VQA (Task B)} 
& LLaVA-1.5~\cite{liu2024improved} & 0.124 & 0.030 & 0.012 & 0.006 & 0.039 & 0.103 & 0.047\\
\cmidrule{2-9}
&\emph{w/o ThinkAloud} & 0.329 & 0.222 & 0.144 & 0.095 & 0.147 & 0.286 & 0.263
 \\
% &\emph{w/o GenAttStrat} & 0.369 & 0.261 & 0.184 & 0.138 & 0.156 & 0.327 & 0.278 \\
% &\emph{w/o BasicComp} & \textbf{0.375} & \textbf{0.278} & \textbf{0.206} & \textbf{0.159} & \textbf{0.169} & \textbf{0.342} & \textbf{0.433}
%  \\
&\emph{Gazette} & \textbf{0.364} & \textbf{0.268} & \textbf{0.202} & \textbf{0.159} & \textbf{0.160} & \textbf{0.324} &\textbf{ 0.367}\\
& & {\scriptsize (+10.64\%)} & {\scriptsize (+20.72\%)} & {\scriptsize (+40.28\%)} & {\scriptsize (+67.37\%)} & {\scriptsize (+8.84\%)} & {\scriptsize (+13.29\%)} & {\scriptsize (+39.54\%)}

 \\

\bottomrule
% \multicolumn{8}{c}{\emph{(b) Decoding VQA (AiR-D)}} \\
\end{tabular}

%  \setlength{\tabcolsep}{3pt}
% \begin{tabular}{llccccccccc}
% % \hline

% \hline
% \textbf{Task} & \textbf{Method} & \textbf{B-1} & \textbf{B-2} & \textbf{B-3} & \textbf{B-4} & \textbf{M} & \textbf{R-L} & \textbf{C} \\
% \hline\addlinespace[2pt]
% \multirow{3}{1cm}{Object Referral} & RefCOCO inter-annotator consistency & 0.500 & 0.282 & 0.148 & 0.057 & 0.244 & 0.452 & 1.115\\
% \cmidrule{2-9}
%  & \emph{w/o ThinkAloud} & 0.479 & 0.263 & 0.145 & \underline{0.070} & 0.232 & 0.443 & 0.872 \\
% % & \emph{w/o GenAttStrat} & 0.493 & \underline{0.296} & \underline{0.153} & \underline{0.085} & 0.243 & \underline{0.458} & 0.937 \\
% % &\emph{w/o BasicComp} & 0.472 & 0.259 & 0.134 & \underline{0.061} & 0.227 & 0.426 & 0.838 \\
% &\emph{Gazette} & \underline{\textbf{0.519}} & \underline{\textbf{0.305}} & \underline{\textbf{0.175}} & \underline{\textbf{0.098}} & \underline{\textbf{0.248}} & \underline{\textbf{0.480}} & \textbf{0.974}
%  \\\addlinespace[2pt]
% \hline\addlinespace[2pt]

% \multirow{2}{*}{VQA} &\emph{w/o ThinkAloud} & 0.329 & 0.222 & 0.144 & 0.095 & 0.147 & 0.286 & 0.263
%  \\
% % &\emph{w/o GenAttStrat} & 0.369 & 0.261 & 0.184 & 0.138 & 0.156 & 0.327 & 0.278 \\
% % &\emph{w/o BasicComp} & \textbf{0.375} & \textbf{0.278} & \textbf{0.206} & \textbf{0.159} & \textbf{0.169} & \textbf{0.342} & \textbf{0.433}
% %  \\
% &\emph{Gazette} & \textbf{0.364} & \textbf{0.268} & \textbf{0.202} & \textbf{0.159} & \textbf{0.160} & \textbf{0.324} &\textbf{ 0.367}
%  \\

% \hline
% % \multicolumn{8}{c}{\emph{(b) Decoding VQA (AiR-D)}} \\
% \end{tabular}
\end{table*}

% \item \emph{Visual Question Answering (VQA)}
% \begin{enumerate}
%     \item \textbf{Question Overlap} - How much does the generated question match the ground truth question, especially in terms of coverage of the entities, their attributes and spatial relationships?

%     \item \textbf{Answer Equivalence} - Does the generated question and the ground-truth question have the same correct answer on the image?
% \end{enumerate}
% \end{enumerate}

 %Primarily, we assess 

As shown in Table~\ref{tab:lexical_res}, Gazette trained with the auxiliary \emph{ThinkAloud} task significantly outperforms the variant  not trained on \emph{ThinkAloud} (``\emph{w/o ThinkAloud}''), and baselines LLaVA-1.5~\cite{liu2024improved} and \emph{LLaVA-last} for both Task A and Task B on every lexical overlap metric. Under the LLM-as-a-Judge paradigm~(Table~\ref{tab:gpt4_eval}), the reported scores follow the same trend seen in lexical overlap metrics, with Gazette outperforming the variant not trained on \emph{ThinkAloud} (``\emph{w/o ThinkAloud}'') by large margins on every human-interpretable rubric. This shows the efficacy of deeper reasoning about the top-down attentional processes via the \emph{ThinkAloud} task. \emph{LLaVA-last} baseline performs poorly in Tables~\ref{tab:lexical_res} and \ref{tab:gpt4_eval}, suggesting that complete scanpaths provide crucial context for referring expression generation. In the supplement, we explore the individual effects of idea units within think-aloud transcripts, and show that for Object Referral, learning both attention allocation explanation generation and target localization enhances performance. For VQA  (where target localization is not relevant), only attention allocation explanation generation is crucial, as it helps Gazette learn complex reasoning processes underlying the visual question answering task.
%Gazette even exceeds the RefCOCO inter-annotator consistency values for almost every metric, suggesting that it effectively covers multiple attributes and properties of the referred object. On close inspection, we observe that excluding \emph{BasicComp} affects performance more than \emph{GenAttStrat}. In the supplementary material, we show that within \emph{BasicComp}, excluding \emph{BasicComp\textsubscript{loc}} affects these metrics more than excluding \emph{BasicComp\textsubscript{count}}, suggesting that localization is closely related to gaze behavior for referral. However, only when  \emph{BasicComp} is combined with  \emph{GenAttStrat} to yield our proposed Gazette model, do we achieve the best performance for Referral  gaze decoding. On the other hand, \emph{GenAttStrat} is more crucial for improved performance for Gazette on the VQA task, and exclusion of \emph{BasicComp} enables Gazette to focus on learning the complex reasoning processes underlying VQA. This is expected since target localization is relevant for search and object referral, not VQA. 

\begin{table*}[ht!]
\centering
\caption{Performance of Gazette and baselines on Task A -- Object Referral Gaze Decoding evaluated using RefCOCO-Gaze dataset~\cite{mondal2025look}, and Task B -- VQA Gaze Decoding evaluated using AiR-D dataset~\cite{chen2020air} by GPT-4~\cite{achiam2023gpt} on a scale of 1-10 under the LLM-as-a-Judge setting. Scores are averaged with best scores in bold.}
\label{tab:gpt4_eval}
\begin{tabular}{llccc}
\toprule
&&\multicolumn{2}{c}{\textbf{Method}}\\
\cmidrule{3-5}
Task & Rubric & \emph{LLaVA-last} &\emph{w/o ThinkAloud} & \emph{Gazette}\\

\midrule
% \multirow{3}{1cm}{Object Referral}
%  & Expression Overlap & 4.697  & \textbf{5.646} \\
%  & Referential Equivalence & 8.548 & \textbf{8.363}
%  \\
%  & Category Correctness & 5.756 & \textbf{6.681}
%  \\
% \midrule
% \multirow{2}{1cm}{VQA}
%  & Question Overlap & 2.103  & \textbf{2.792}
%  \\
%  & Answer Equivalence & 1.642  & \textbf{2.135} \\
\multirow{3}{1.5cm}{Object Referral (Task A)}
 & Expression Overlap & 3.515  & 5.019 & \textbf{5.980} \\
 & Referential Equivalence & 4.100 & 5.771 & \textbf{6.638}
 \\
 & Category Correctness & 7.577 & 8.376 & \textbf{8.743}
 \\
\midrule
\multirow{2}{1.5cm}{VQA (Task B)}
 & Question Overlap & - & 2.103  & \textbf{2.792}
 \\
 & Answer Equivalence & - & 1.642  & \textbf{2.135} \\
\bottomrule
\end{tabular}
% \begin{tabular}{llcccc}
% \toprule
% &&\multicolumn{4}{c}{\textbf{Method}}\\
% \cmidrule{3-6}
% & & \multirow{3}{2cm}{\centering \emph{w/o BasicComp \& GenAttStrat}} & \multirow{3}{2cm}{\centering \emph{w/o GenAttStrat}} & \multirow{3}{2cm}{\centering \emph{w/o BasicComp}} & \multirow{3}{2cm}{\centering \emph{Gazette}}\\
% & & & & &\\

% Task & Rubric & & & & \\
% \midrule
% \multirow{3}{1cm}{Object Referral}
%  & Expression Overlap & 4.697 & 5.368 & 5.378 & \textbf{5.646} \\
%  & Referential Equivalence & 8.548 & 8.469 & 8.723 & \textbf{8.363}
%  \\
%  & Category Correctness & 5.756 & 6.389 & 6.318 & \textbf{6.681}
%  \\
% \midrule
% \multirow{2}{1cm}{VQA}
%  & Question Overlap & 2.103 & 2.279 & 2.668 & \textbf{2.792}
%  \\
%  & Answer Equivalence & 1.642 & 1.707 & \textbf{2.173} & 2.135 \\
% \bottomrule
% \end{tabular}
\end{table*}

\begin{table*}[ht!]
\centering
\caption{Performance of Gazette and baselines on Task C (Target-Present) and Task D (Target-Absent) Visual Search Gaze Decoding, both evaluated using COCO-Search18 dataset~\cite{chen2021coco}. Best results are in bold, and second best are underlined.}
\label{tab:visual_search_res}
\begin{tabular}{llcccc}
% \hline

\toprule 
\textbf{Task} & \textbf{Method} & \textbf{Precision} & \textbf{Recall} & \textbf{F\textsubscript{1}} & \textbf{Accuracy} \\
\hline\addlinespace[2pt]
 \multirow{7}{1.5cm}{Target-Present Visual Search (Task C) }
% &\emph{w/o GenAttStrat} & 0.765 & 0.745 & 0.745 & 0.769 \\
% &\emph{w/o BasicComp} & 0.692 & 0.686 & 0.679 &0.686 \\
&LLaVA-1.5~\cite{liu2024improved} & 0.448 & 0.436 & 0.406 & 0.402\\
&\emph{LLaVA-last} & 0.688 & 0.649 & 0.628 & 0.618\\
&GazeGNN~\cite{wang2024gazegnn} & 0.338 & 0.345 & 0.319 & 0.335\\
&GST~\cite{nishiyasu2024gaze} & - & - & - & 0.544\\
&Mondal~\etal~\cite{Mondal_2025_ICCV} & - & - & - & \textbf{0.776} \\
\cmidrule{2-6}

& \emph{w/o ThinkAloud} & \underline{0.768} & \underline{0.746} & \underline{0.742} & 0.742 \\

&\emph{Gazette} & \textbf{0.786} & \textbf{0.775} & \textbf{0.775} & \underline{0.773}\\

\hline\addlinespace[2pt]
% 0.33790014408829167, 0.3450837395035291, 0.3193777722931628
\multirow{6}{1.5cm}{Target-Absent Visual Search (Task D)}
&LLaVA-1.5~\cite{liu2024improved} & 0.128 & 0.102 & 0.077 & 0.088\\
&GazeGNN~\cite{wang2024gazegnn} & 0.241 & 0.251 & 0.218 & 0.270\\
&GST~\cite{nishiyasu2024gaze} & - & - & - & 0.385\\
&Mondal~\etal~\cite{Mondal_2025_ICCV} & - & - & - & 0.387\\
\cmidrule{2-6}

& \emph{w/o ThinkAloud} & \underline{0.437} & \underline{0.424} & \underline{0.420} & \textbf{0.445} \\
% &\emph{w/o GenAttStrat}  & \textbf{0.445} & \textbf{0.426} & \textbf{0.424} & \textbf{0.456}\\
% &\emph{w/o BasicComp} & 0.422 & 0.418 & 0.412 &0.438 \\
&\emph{Gazette} & \textbf{0.438} & \textbf{0.426} & \textbf{0.424} & \underline{0.443}\\

\bottomrule
% \multicolumn{8}{c}{\emph{(b) Decoding VQA (AiR-D)}} \\
\end{tabular}

\end{table*} 
 \def\subFigSz{\linewidth}
 \begin{figure*}[ht!]
\centering
\makebox[\subFigSz][h]{\small{Gazette Predicts: Object Referral for \textbf{``black car top right''}.}}\\
\makebox[\subFigSz][h]{\small{Gazette w/o ThinkAloud Predicts: Object Referral for \textbf{``red car''}.}}

%tp-000000534121.jpg-tv-9
\makebox[\subFigSz][h]{\includegraphics[width=\linewidth]{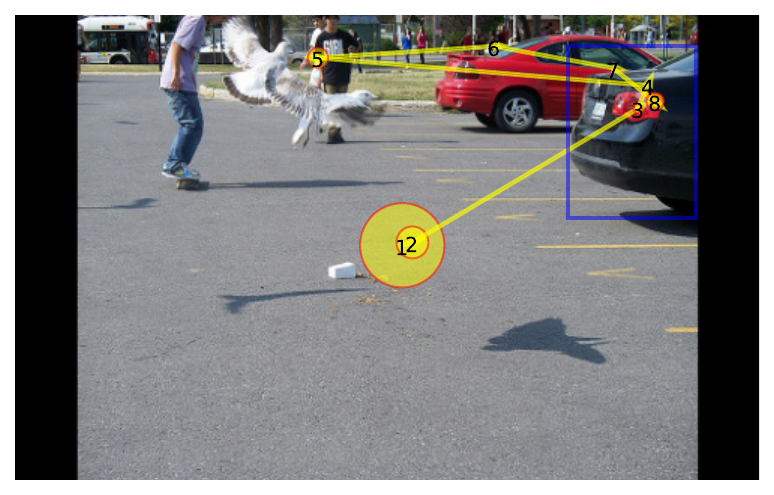} }
\makebox[\subFigSz][h]{\begin{minipage}[t]{\linewidth}
      \small{\textbf{Gazette-generated explanation}:} \small{Most humans initially fixate on the road before shifting their attention to the car located in the upper right corner of the image, often after scanning other cars and occasionally focusing on people in the vicinity.
      }
    \end{minipage}}
    \caption{\textbf{Qualitative Results.} Comparison of Gazette variants on decoding a gaze scanpath for \textbf{Object Referral} target \textbf{``black car on right''} (blue bounding box). We show predictions from Gazette and its variant not trained on \emph{ThinkAloud}, along with the attention allocation explanation from the Gazette-generated think-aloud transcript.} 
% \vskip -0.1in
\label{fig:qualitative_ref}
\end{figure*}

Next, we focus on evaluating the target prediction capabilities of Gazette (trained with and without \emph{ThinkAloud}), and compare with baselines GazeGNN~\cite{wang2024gazegnn} (which we adapt for target prediction), GST~\cite{nishiyasu2024gaze}, Mondal \etal~\cite{Mondal_2025_ICCV}, LLaVA-1.5~\cite{liu2024improved}, and \emph{LLaVA-last}, on the COCO-Search18 benchmark~\cite{chen2021coco}. The results are shown in Table~\ref{tab:visual_search_res}. The implementations for GST and Mondal \etal are not publicly available, therefore we are unable to furnish metrics not reported by the authors (\ie precision, recall, $F_1$). Gazette significantly outperforms prior methods across all metrics on the Target-Absent Visual Search task, and achieves accuracy in the Target-Present setting that is nearly on par with the state-of-the-art visual search gaze decoding method from Mondal \etal (0.773 vs. 0.776). \emph{LLaVA-last} fares well in recognizing Target-Present gaze targets, but still lags behind Gazette, suggesting the importance of scanpath context. In the supplement, we show that similar to Object Referral, both target localization and attention allocation explanation generation idea units are important for Target-Present search gaze decoding. For Target-Absent trials, training with \emph{ThinkAloud} results in modest and mixed performance improvements. We attribute this to poor agreement between observers in Target-Absent search~\cite{yang2022target}, where gaze behavior increasingly resembles free-viewing as search progresses~\cite{chen2022characterizing}, resulting in low commonality across participants and weak shared attentional patterns — potentially degrading GPT-4's responses when prompted by our strategy. We show in the supplement that predicting the absence of the target and measuring the length of the scanpath improves
Target-Absent search gaze decoding.

Fig.~\ref{fig:qualitative_ref} qualitatively analyzes generated texts from Gazette trained either with or without \emph{ThinkAloud} instructions to decode a scanpath of a human grounding the referring expression ``black car on right''. Without a deeper understanding of the verification and scanning strategies exhibited by the scanpath through \emph{ThinkAloud} instruction tuning, the model misinterprets the ``red car'' to be the target, whereas our full Gazette model trained on \emph{ThinkAloud} instructions identifies the correct car by analyzing the gaze patterns, as evidenced by the attention allocation explanation text generated by Gazette. In the supplement, we qualitatively analyze additional samples of model-decoded cognitive contexts and  think-aloud transcripts generated by Gazette. %However, we show in the supplementary material that predicting the absence of the target (embodied by \emph{BasicComp\textsubscript{loc}}) proves to be more useful. We show in the supplementary material that \emph{BasicComp\textsubscript{loc}} is more important than \emph{BasicComp\textsubscript{count}} for both Target Present and Target Absent Visual Search tasks.

\section{Conclusion}
\label{sec:conclusion}

In this work, we explored the capabilities of Multimodal LLMs in understanding human attention behavior, specifically top-down attention control. We proposed \emph{Gazette}, a novel text-generative gaze decoding framework to decode a wide array of top-down attention behaviors. Rather than naively instruction-tuning MLLMs on the primary gaze decoding task, we built an auxiliary dataset for generating top-down attentional allocation explanations -- termed \emph{think-aloud transcripts} -- to encourage Gazette to explicitly separate goal-specific attentional dynamics from individual idiosyncrasies. To generate pseudo-annotations for these think-aloud transcripts, we prompted GPT-4 using a novel prompting strategy which exploits commonalities in gaze patterns across participants engaged in the same attentional task for the same image. We showed that when trained with a combination of our proposed primary and auxiliary tasks, Gazette achieved significant performance boost in both generative and predictive gaze decoding tasks over naive instruction tuning strategies and state-of-the-art methods, as evaluated by our rigorous evaluation scheme. Gazette opens the door to applications that hinge on decoding precisely what a person is looking for, down to fine-grained details, bringing such capabilities within the reach of real-world systems. While we demonstrated generalization across a diverse set of top-down attentional behaviors, Gazette can potentially be extended to psychological analysis, driving, and assistive healthcare—domains that require precise decoding of human mental states. We expect that Gazette, together with our novel GPT-4 prompting strategy for generating think-aloud transcript pseudo-annotations, will inspire future work exploring new avenues of gaze understanding with LLMs and MLLMs.

\paragraph{\textbf{Limitations.}} Our method is not without limitations. Its reliance on inter-observer consistency to extract top-down attentional control patterns means efficacy degrades on tasks with low observer agreement, \eg Target-Absent Visual Search. In addition, because the think-aloud transcripts are GPT-4-generated, they inherit its training priors rather than participants' true cognition, raising a concern of \emph{bias} toward a majority-weighted notion of attention allocation. We leave addressing these limitations to future work. 

\paragraph{\textbf{Acknowledgements.}} This project was supported by US National Science Foundation Award IIS-1763981, IIS-2123920, DUE-2055406, and the SUNY2020 Infrastructure
Transportation Security Center, and a gift from Adobe. We are grateful to Abe Leite for his invaluable help with the human evaluation of the GPT-4-generated top-down attention allocation explanation pseudo-annotations, to Niranjan Balasubramanian for his generous guidance and discussion, and to Xiang Li for his crucial help with platform issues during MLLM training.

\clearpage  % TODO FINAL: This \clearpage needs to be removed from both review and camera-ready versions.

% \section*{Acknowledgements}
% Please insert your acknowledgments here.


\chapter*{Supplementary Material}
\setcounter{section}{0}
\renewcommand{\thesection}{\Roman{section}}

In this document, we provide in-depth details, ablations, and qualitative analysis of our gaze-to-text generative framework \emph{Gazette}, along with specifics of our evaluation method and baselines. The sections of this document are detailed below: 

\begin{enumerate}
    \item In Sec.~\ref{sec:idea_units}, we probe the effects that the different ``idea units'' in think-aloud transcripts have on gaze decoding performance.
    \item In Sec.~\ref{sec:behavior_type}, we discuss the decoding capabilities of our model on predicting the behavior type facet of the cognitive context. 

    \item In Sec.~\ref{sec:prompt_appendix}, we detail the prompt templates used for prompting Gazette and GPT-4 in our paper, and also present results and analyses for human evaluation of the quality of GPT-4-generated attention allocation explanation pseudo-annotations when prompted by our novel prompting strategy.

    \item In Sec.~\ref{sec:impl}, we provide in-depth implementation details for Gazette and the several baselines used in our paper.

    \item In Sec.~\ref{sec:qual}, we provide qualitative examples of Gazette, where we juxtapose model predictions with generated attention allocation explanation idea unit texts.
\end{enumerate}
% \clearpage  % TODO FINAL: This \clearpage needs to be removed from both review and camera-ready versions.
\section{Probing Idea Units in Think-Aloud Transcripts}
\label{sec:idea_units}

In Sec. 3.2 of the main text, we mentioned that a think-aloud transcript contains three ``idea units''. In this section, we will discuss their impacts on gaze decoding. Specifically, these three idea units are: (i) top-down attention allocation explanation, which we denote as \emph{GenAttAlloc}, (ii) target location, which we denote as \emph{LocTarget}, and (iii) scanpath length, which we denote as \emph{CountFix}. We probe the effects of these idea units on performance through a series of ablations on the four gaze behaviors in our studies. In each ablation, we either remove or retain each idea unit in the think-aloud transcript. Note that for each ablation, $\mathbf{T}_{ThinkAloud}$ is identical for fair comparison. The results are in Table~\ref{tab:ideaunit_ref}, Table~\ref{tab:ideaunit_vqa}, Table~\ref{tab:ideaunit_tp}, and Table~\ref{tab:ideaunit_ta}, for Object Referral, Visual Question Answering (VQA), Target-Present Visual Search, and Target-Absent Visual Search, respectively.

In Table~\ref{tab:ideaunit_ref}, we observe that for Object Referral, while \emph{GenAttAlloc} idea unit (\ie attention allocation explanation generation sub-task) is crucial, only when it is combined with the target localization sub-task, do we achieve the best performance. We also note that \emph{CountFix} and \emph{LocTarget} are not sufficient for Gazette to learn how to decode object referral gaze optimally. A similar trend is seen in Table~\ref{tab:ideaunit_tp} for Target-Present Search, where localization is more important than any other idea unit. We attribute this to short scanpaths for Target-Present Visual Search, where fixations usually land on the target within 1-2 fixations after the initial fixation. On the other hand, \emph{GenAttAlloc} embodying the attention allocation explanation generation task is most crucial for VQA Gaze Decoding (see Table~\ref{tab:ideaunit_vqa}), with other idea units negatively affecting performance when added  to the think-aloud transcript (see row 5 in Table~\ref{tab:ideaunit_vqa}). However, for Target-Absent Visual Search (Table~\ref{tab:ideaunit_ta}), predicting the absence of the target (embodied by \emph{LocTarget}) and counting fixations in a scanpath (\emph{CountFix}) seem to improve performance more than \emph{GenAttAlloc} does. We attribute this to low commonality in gaze patterns across participants for Target-Absent Search, potentially affecting GPT’s responses when prompted by our strategy. Overall, we find a combination of all idea units is most beneficial, prompting us to include all of them in the think-aloud transcript annotations.

\begin{table*}[ht!]
\setlength{\tabcolsep}{1pt}
\centering
\caption{Probing effects of idea units (\emph{GenAttAlloc, LocTarget, CountFix}) in think-aloud transcripts on Object Referral Gaze Decoding. Best results are highlighted in bold. The last row signifies our full model, \emph{Gazette}.}
\label{tab:ideaunit_ref}
\begin{tabular}{cccccccccc}
% \hline

\toprule 
% &  & \multicolumn{4}{c}{\textbf{\small BLEU-n}}   &  \multirow{2}{*}{\scriptsize METEOR} & \\
%  \cmidrule{3-6}
 \emph{GenAttAlloc} &  \emph{LocTarget} &  \emph{CountFix} &{\bf \scriptsize BLEU-1} & {\bf \scriptsize BLEU-2} & {\bf \scriptsize BLEU-3} & {\bf \scriptsize BLEU-4} & {\bf \scriptsize METEOR} & \textbf{\scriptsize ROUGE-L} & \textbf{\scriptsize CIDEr} \\
%\hline\addlinespace[2pt]
\midrule 
$\times$ & $\times$ & $\times$  & 0.479 & 0.263 & 0.145 & 0.070 & 0.232 & 0.443 & 0.872 \\
$\checkmark$  & $\times$ & $\checkmark$ & 0.455 & 0.241 & 0.121 & 0.061 & 0.224 & 0.429 & 0.806\\
 $\checkmark$  & $\checkmark$ & $\times$ & 0.485 & 0.275 & 0.152 & 0.076 & 0.233 & 0.452 & 0.910\\
 $\times$ & $\checkmark$  & $\checkmark$  & 0.494 & 0.272 & 0.153 & 0.085 & 0.239 & 0.446 & 0.918\\
% $\times$& $\checkmark$ & $\checkmark$ & 0.493 & 0.296 & 0.153 & 0.085 & 0.243 & 0.458 & 0.937 \\
$\checkmark$& $\times$& $\times$& 0.472 & 0.259 & 0.134 & 0.061 & 0.227 & 0.426 & 0.838 \\
$\checkmark$& $\checkmark$& $\checkmark$& \textbf{0.519} & \textbf{0.305} & \textbf{0.175} & \textbf{0.098} & \textbf{0.248} & \textbf{0.480} & \textbf{0.974}\\
\bottomrule
% \multicolumn{8}{c}{\emph{(b) Decoding VQA (AiR-D)}} \\
\end{tabular}
\end{table*}

\begin{table*}[ht!]
\centering
\caption{Probing effects of idea units (\emph{GenAttAlloc, LocTarget, CountFix}) in think-aloud transcripts on VQA Gaze Decoding. Best results are highlighted in bold. The last row signifies our full model, \emph{Gazette}.}
\label{tab:ideaunit_vqa}
\setlength{\tabcolsep}{1pt}

\begin{tabular}{cccccccccccc}
% \hline

\toprule 
% &  & \multicolumn{4}{c}{\textbf{\small BLEU-n}}   &  \multirow{2}{*}{\scriptsize METEOR} & \\
%  \cmidrule{3-6}
 \emph{GenAttAlloc} &  \emph{LocTarget} &  \emph{CountFix} &{\bf \scriptsize BLEU-1} & {\bf \scriptsize BLEU-2} & {\bf \scriptsize BLEU-3} & {\bf \scriptsize BLEU-4} & {\bf \scriptsize METEOR} & \textbf{\scriptsize ROUGE-L} & \textbf{\scriptsize CIDEr} \\
%\hline\addlinespace[2pt]
\midrule 
 $\times$& $\times$& $\times$& 0.329 & 0.222 & 0.144 & 0.095 & 0.147 & 0.286 & 0.263 \\
$\checkmark$  & $\times$ & $\checkmark$  & 0.354 & 0.252 & 0.178 & 0.133 & 0.155 & 0.316 & 0.300\\
 
 $\checkmark$  & $\checkmark$ & $\times$  & 0.364 & 0.264 & 0.193 & 0.147 & 0.163 & 0.330 & 0.310\\

$\times$& $\checkmark$& $\checkmark$& 0.369 & 0.261 & 0.184 & 0.138 & 0.156 & 0.327 & 0.278  \\
$\checkmark$& $\times$ & $\times$ & \textbf{0.375} & \textbf{0.278} & \textbf{0.206} & \textbf{0.159} & \textbf{0.169} & \textbf{0.342} & \textbf{0.433} \\
$\checkmark$& $\checkmark$& $\checkmark$&  0.364 & 0.268 & 0.202 & \textbf{0.159} & 0.160 & 0.324 &0.367\\
\bottomrule

\end{tabular}
\end{table*}

\begin{table*}[ht!]
\centering
\caption{Probing effects of idea units (\emph{GenAttAlloc, LocTarget, CountFix}) in think-aloud transcripts on Target-Present Visual Search Gaze Decoding. Best results are highlighted in bold. The last row signifies our full model, \emph{Gazette}.}
\label{tab:ideaunit_tp}
\begin{tabular}{cccccccc}
% \hline

\toprule 
% &  & \multicolumn{4}{c}{\textbf{\small BLEU-n}}   &  \multirow{2}{*}{\scriptsize METEOR} & \\
%  \cmidrule{3-6}
 \emph{GenAttAlloc} &  \emph{LocTarget} &  \emph{CountFix} &{\bf Precision} & {\bf Recall} & {\bf F$_1$} & {\bf Accuracy} \\
%\hline\addlinespace[2pt]
\midrule 
 $\times$& $\times$& $\times$& 0.768 & 0.746 & 0.742 & 0.742 \\
$\checkmark$  & $\times$ & $\checkmark$  & 0.714 & 0.709 & 0.701 & 0.703\\
 
 $\checkmark$  & $\checkmark$ & $\times$  & 0.759 & 0.753 & 0.749 & 0.748\\

$\times$& $\checkmark$& $\checkmark$& 0.765 & 0.745 & 0.745 & 0.749  \\
$\checkmark$& $\times$ & $\times$ & 0.692 & 0.686 & 0.679 & 0.686 \\
$\checkmark$& $\checkmark$& $\checkmark$&  \textbf{0.786} & \textbf{0.775} & \textbf{0.775} & \textbf{0.773}\\
\bottomrule

\end{tabular}
\end{table*}

\begin{table*}[ht]
\centering
\caption{Probing effects of idea units (\emph{GenAttAlloc, LocTarget, CountFix}) in think-aloud transcripts on Target-Absent Visual Search Gaze Decoding. Best results are highlighted in bold. The last row signifies our full model, \emph{Gazette}.}
\label{tab:ideaunit_ta}
\begin{tabular}{cccccccc}
% \hline

\toprule 
% &  & \multicolumn{4}{c}{\textbf{\small BLEU-n}}   &  \multirow{2}{*}{\scriptsize METEOR} & \\
%  \cmidrule{3-6}
 \emph{GenAttAlloc} &  \emph{LocTarget} &  \emph{CountFix} &{\bf Precision} & {\bf Recall} & {\bf F$_1$} & {\bf Accuracy} \\
%\hline\addlinespace[2pt]
\midrule 
 $\times$& $\times$& $\times$& 0.437 & 0.424 & 0.420 & 0.445\\
$\checkmark$  & $\times$ & $\checkmark$  & 0.421 & 0.409 & 0.409 & 0.427\\
 
 $\checkmark$  & $\checkmark$ & $\times$  & 0.434 & 0.425 & 0.423 & 0.445\\
$\times$& $\checkmark$& $\checkmark$& \textbf{0.445} & \textbf{0.426} & \textbf{0.424} & \textbf{0.456} \\
$\checkmark$& $\times$ & $\times$ & 0.422 & 0.418 & 0.412 & 0.438 \\
$\checkmark$& $\checkmark$& $\checkmark$&  0.438 & \textbf{0.426} & \textbf{0.424} & 0.443\\
\bottomrule

\end{tabular}

\end{table*}

\begin{table*}[ht!]
\centering
\caption{Performance of Gazette (full model and w/o \emph{ThinkAloud}) on gaze behavior type prediction as measured by precision, recall, and F$_1$ scores.}
\label{tab:behavior_type}
\begin{tabular}{llccc}
% \hline

\toprule 
\textbf{Task} & \textbf{Method} & \textbf{Precision} & \textbf{Recall} & \textbf{F\textsubscript{1}} \\
\hline\addlinespace[2pt]
 \multirow{2}{*}{Object Referral (Task A) }
& \emph{w/o ThinkAloud} & 0.99 & 0.95 & 0.97 \\

&\emph{Gazette} & 0.93 & 0.99 & 0.96\\

\hline\addlinespace[2pt]
% 0.33790014408829167, 0.3450837395035291, 0.3193777722931628
\multirow{2}{*}{Visual Question Answering (Task B)}
& \emph{w/o ThinkAloud} & 0.99 & 1.00 & 1.00  \\
% &\emph{w/o GenAttAlloc}  & \textbf{0.445} & \textbf{0.426} & \textbf{0.424} & \textbf{0.456}\\
% &\emph{w/o BasicComp} & 0.422 & 0.418 & 0.412 &0.438 \\
&\emph{Gazette}  & 1.00 & 0.98 & 0.99  \\
\hline\addlinespace[2pt]
 \multirow{2}{*}{Target-Present Visual Search (Task C) }
& \emph{w/o ThinkAloud} & 0.84 & 0.81 & 0.82 \\

&\emph{Gazette} & 0.85 & 0.83 & 0.84\\

\hline\addlinespace[2pt]
% 0.33790014408829167, 0.3450837395035291, 0.3193777722931628
\multirow{2}{*}{Target-Absent Visual Search (Task D)}
& \emph{w/o ThinkAloud} & 0.81 & 0.85 & 0.83  \\
% &\emph{w/o GenAttAlloc}  & \textbf{0.445} & \textbf{0.426} & \textbf{0.424} & \textbf{0.456}\\
% &\emph{w/o BasicComp} & 0.422 & 0.418 & 0.412 &0.438 \\
&\emph{Gazette}  & 0.83 & 0.84 & 0.84  \\
\bottomrule
\multirow{2}{*}{Tasks A, B, C, and D combined} & \emph{w/o ThinkAloud} & 0.91 & 0.90 & 0.90  \\
&\emph{Gazette}  & 0.90 & 0.91 & 0.91  \\
% \multicolumn{8}{c}{\emph{(b) Decoding VQA (AiR-D)}} \\
\bottomrule
\end{tabular}

\end{table*} 

\section{Gaze Behavior Type Prediction Results}
\label{sec:behavior_type}

The gaze behavior type prediction results for Gazette with and without \emph{ThinkAloud} instruction tuning are provided in Table~\ref{tab:behavior_type}. 
As mentioned in Sec.4 of the main text, we find that gaze behavior type prediction is trivial for Gazette, achieving more than 0.90 across precision, recall, and F$_1$ scores, regardless of whether it is trained with or without \emph{ThinkAloud} instructions. We attribute this to the fact that the four gaze behaviors we study are derived from three distinct datasets, collected under different behavioral setups, potentially contributing to artifacts that can easily be recognized by the model without additional supervision from think-aloud transcripts. However, following previous gaze decoding work~\cite{nishiyasu2024gaze} and for the sake of completeness, we retain the coarser behavior type facet of the cognitive context. The main focus of the paper remains decoding the finer stimulus facet of the cognitive context, which has been shown to be a non-trivial task in this paper.

\section{Prompting procedures for Gazette and GPT-4}
\label{sec:prompt_appendix}

\subsection{Gazette Instruction Tuning.}The instruction tuning data used to fine-tune Gazette is a combination of two types of instructions for the same scanpath-image input, where the scanpath is generally denoted as a series of fixations (x,y,t): [($x_0, y_0, t_0$), ..., ($x_{N-1}, y_{N-1}, t_{N-1}$)]. Note that the spatial coordinates x and y are normalized while fixation duration t is unnormalized. An example fixation is: (0.661, 0.112, 272.0).

\begin{itemize}
    \item \textbf{Instruction $\mathbf{T}_{GazeDec}$ for primary task \emph{GazeDec}:}

We use the following prompt template to construct instruction $\mathbf{T}_{GazeDec}$ for primary task \emph{GazeDec}: 
\begin{lstlisting}
    Given a scanpath [($x_0, y_0, t_0$), ..., ($x_{N-1}, y_{N-1}, t_{N-1}$)]\texttt{, which is a list of fixations [x, y, t], with spatial co-ordinates x and y normalized between 0 and 1 and t as the fixation duration), analyze the sequence in the context of the given image to infer the underlying cognitive process defined by the top-down task \#TASK and the top-down stimulus \#STIMULUS. Approach the problem through these systematic steps: (1) Identify the task type (Target-Present Search, Target-Absent Search, Object Referral, or Visual Question Answering) and label it as \#TASK. (2) Generate a description of the top-down stimulus or goal, marking it as \#STIMULUS. (3) Format your final response as: <task> \#TASK </task> <stimulus> \#STIMULUS </stimulus>}
\end{lstlisting}
The ground truth response is in XML format for easy parsing of the textual response from Gazette to $\mathbf{D}_{type}$ and $\mathbf{D}_{goal}$. A sample response is: \texttt{<task> Object Referral </task> <stimulus> red car on the left </stimulus>}
\\\\\\

 \item \textbf{Instruction $\mathbf{T}_{ThinkAloud}$ for \emph{ThinkAloud}}
 
We follow recent instruction tuning work~\cite{li2024mosaic, cai2025mergeit} and combine all idea units of the think-aloud transcript into a single instruction $\mathbf{T}_{ThinkAloud}$, thus avoiding training Gazette on the same image-scanpath pair multiple times that leads to poor generalization on our limited gaze training data. The prompt template for $\mathbf{T}_{ThinkAloud}$ is the following: \\

\begin{lstlisting}
    Given a scanpath }[($x_0, y_0, t_0$), ..., ($x_{N-1}, y_{N-1}, t_{N-1}$)]\texttt{, which is a list of fixations [x, y, t], with spatial co-ordinates x and y normalized between 0 and 1 and t as the fixation duration), analyze the sequence in the context of the given image to describe the attention allocation strategy \#STRATEGY. Approach the problem through these systematic steps: (1) Describe the attentive strategy based on fixation density, areas of interest, durations, and spatiotemporal patterns between fixations and label this text as \#STRATEGY. (2) Generate \#STRATEGY as: <strategy> \#STRATEGY </strategy>"
\end{lstlisting}

\end{itemize}

\def\subFigSz{\linewidth}
% \begin{figure*}[ht!]
%   \centering
%   \makebox[0.42\subFigSz][h]{\small{Stimulus: \textbf{Object Referral} for \textbf{``car on the right''}.}}\\
%  \makebox[0.42\subFigSz][h]{ \includegraphics[width=0.42\linewidth]{figures/rgo1210983.png}}
%  \makebox[0.42\subFigSz][h]{ \includegraphics[width=0.42\linewidth]{figures/rgo1215121.png}}\\
%  \makebox[0.42\subFigSz][h]{ \includegraphics[width=0.42\linewidth]{figures/rgo1217229.png}}
%  \makebox[0.42\subFigSz][h]{ \includegraphics[width=0.42\linewidth]{figures/rgo1225672.png}}\\
% \begin{minipage}[t]{\linewidth}{\small{GPT-4 generated: \textbf{Most humans initially fixate on the large vertical parking meter spanning a significant portion of the left-central part of the image, then shift focus toward the large car located on the right side, spending varying durations there.}}}
% \end{minipage}\\
%   \caption{For the given image and a give top-down task (object referral for referring expression \textbf{``car on the right"}, GPT-4 summarizes the common spatiotemporal patterns across several scanpaths (here, we show four scanpaths for the same image-stimulus pair) as per our novel prompting strategy. }
%   \label{fig:gpt4_sample}
% \end{figure*}

\subsection{GPT-4 prompting strategy for generating attention allocation explanation ground truth}
\label{sec:thinkaloud_gpt4}

The template for querying GPT-4 to generate attention allocation explanation using our novel prompting strategy is as follows: 

\textit{System prompt}: 
\begin{lstlisting}
    You are a specialist in human visual attention, adept at analyzing images through bounding box annotations and summarizing fixation patterns.
\end{lstlisting}

\textit{User prompt}: 

\begin{lstlisting}
Given an image described by objects represented as <category: [x, y, w, h]>:[SCENE\_OBJECTS], and 10 humans are searching for a TARGET labeled ``bottle`` that is present in the image. Eye movements of each human are recorded as a scanpath - a sequence of fixations, with each fixation represented as <x-location, y-location , fixation duration, object fixated>: Human 0 - <$fix_0$, ..., $fix_n$>, ..., Human 9 - <$fix_0$, ..., $fix_n$>, identify and concisely merge the most common spatiotemporal patterns across all scanpaths in one sentence. Use clear, unambiguous referring expressions containing attributes and spatial relationships to refer to objects, and avoid referencing individual scanpaths, adding extra information, or formatting the response. 

\end{lstlisting}

 Note that \texttt{[SCENE\_OBJECTS]} are represented by a list of scene objects, each represented by the 
 category of the object and its normalized bounding box, \eg <bottle:[0.634, 0.068, 0.058, 0.321]>, <chair:[0.519, 0.002, 0.27, 0.304]>. Additionally, each i-indexed fixation \texttt{fix\_i} is represented as normalized x,y co-ordinates, raw fixation duration, and category of the fixated object, \eg (0.661, 0.112, 272.0, bottle). As mentioned in Sec. 3.2 of the main text, the category of the fixated object is derived from COCO and Visual Genome annotations. Through this prompt, we leverage the general pattern-understanding capabilities of GPT-4 as documented by several works~\cite{mirchandani2023large, weber2024large}, and do not assume that GPT-4 implicitly understands gaze behavior. Previous work on MLLMs has successfully used GPT-4 for spatial and spatiotemporal synthetic training data generation such as LLaVA~\cite{liu2023visual} , VIGC~\cite{wang2024vigc}, and Zhang \etal~\cite{zhang2024llava}. Following their footsteps, we used GPT-4 to synthesize our own auxiliary instruction tuning data which resulted in better performance, as evidenced by our experimental results. %Sample scanpaths for an image-stimulus pair and the corresponding GPT-4 generated response for our prompting strategy are provided in Figure~\ref{fig:gpt4_sample}.  

\noindent\textbf{Human evaluation of quality of GPT-4 generated transcripts. }To evaluate the quality of the GPT-4 generated transcripts, we recruited three participants to rate 50 GPT-4 generated transcripts. For each image-stimulus pair corresponding to a GPT-4-generated transcript, we first sampled 5 scanpaths included in the prompt to GPT-4 for generating that transcript (positive samples). Five more scanpaths corresponding to other image-stimulus pairs were also sampled (negative samples). Hence we obtain 10 scanpaths in total per image-stimulus pair. Each of these 10 scanpaths is overlaid on the given image and presented to three participants to rate (on a scale of 1-5, 1 being very inconsistent and 5 being very consistent) based on the displayed scanpath's consistency with the GPT-4-generated attention allocation description. This assessment showed that the level of hallucination in these GPT-4 generated transcripts is limited, as participants' ratings for positive samples were statistically significantly higher than for negative samples (p < 0.05), both in terms of ratings of each individual participant and ratings of all three participants combined. On the scale of 1-5, positive samples were rated on average 3.28 (±1.29), and negative samples were rated on average 2.04 (±1.06). We performed paired samples t-test and regression modeling to account for the between-item response biases. For paired samples t-test, T-statistic = 17.662 and p-value < 0.001, which also suggests that there is a strong and statistically significant difference between scores for positive and negative samples, with positive samples scoring higher on average. Similarly, mixed linear regression modeling of the survey scores revealed that positive samples had scores about 1.25 points higher than negative samples on average (recall that the survey scores were on a scale of 1-5, making this difference quite significant), and this effect is large and statistically significant (z-value = 21.128, p < 0.001). Also, group variance (0.092) was small compared to residual variance (1.3024), suggesting that most of the variation is within conditions rather than across items. 
 
\subsection{GPT-4 prompt template for LLM-as-a-Judge evaluation. } We use a comparative prompt comparing multiple methods, as done by Xiong \etal~\cite{xiong2024llava}, by providing the texts generated by them in order to give GPT-4 more context in evaluation. A sample is provided below:

\textit{System Prompt:} 
\begin{lstlisting}
You are a decisive evaluation assistant specializing in analyzing images with bounding box annotations and assessing generated expressions. For each case, compare Output A, Output B, Output C, Output D, and Output E against the provided ground truth. Assign distinct scores unless the outputs are truly equivalent. If you judge one output strictly better on any criterion, assign it at least 2 points higher.
\end{lstlisting}

\newpage
\textit{User Prompt}: 
\begin{lstlisting}
Given an image described by objects represented as <category: [x, y, w, h]>: <bicycle:[0.153, 0.434, 0.517, 0.49]>, <car:[0.111, 0.25, 0.485, 0.347]>, <car:[0.68, 0.222, 0.236, 0.271]>, <bench:[0.085, 0.617, 0.643, 0.37]>, <car:[0.61, 0.257, 0.157, 0.111]>, <car:[0.428, 0.243, 0.256, 0.155]>, <truck:[0.083, 0.187, 0.199, 0.191]> , please compare the following generated referring expressions: Output A: ``the car on the far right``, Output B: ``right side suv``, Output C: ``car on right``, Output D: ``white car to right``, Output E: ``white car on right`` with the ground truth expressions [``benz``, ``silver benz``, ``right car``] referring to the object in bounding box [0.68, 0.222, 0.234, 0.269] based on ``Expression Overlap`` (How much does the generated referring expression match the ground truth referring expressions, especially in terms of coverage of the entities, their attributes and spatial relationships?), ``Referential Equivalence`` (Does either of the ground truth expressions refer to the same object in the image as the generated expression?) ``Category Correctness`` (Is the type or category of the referred object mentioned correctly?). Use a scale from 1 to 10 for each criterion, where 1 is poor and 10 is excellent. Format the response in the form of a dictionary with exactly five top-level keys: ``Output A``, ``Output B``, ``Output C``, ``Output D`` and ``Output E``. Each of those keys maps to a nested dictionary with exactly the following keys - ``Expression Overlap``, ``Category Correctness``, and ``Referential Equivalence`` - whose values are your numeric scores.  Do not include explanations, provide only the scores.
\end{lstlisting}

\section{Implementation Details}
\label{sec:impl}

\noindent\textbf{\emph{Gazette}.} We initialize the weights of Gazette from pre-trained LLaVA-1.5-7B~\cite{liu2024improved} architecture. This model houses a Vicuna~\cite{vicuna2023} v1.5 LLM and, for the vision encoder, uses a CLIP~\cite{radford2021learning} ViT-large image encoder. We train all Gazette variants using LoRA~\cite{hu2021lora} (rank 64) for a maximum of 3 epochs, with a learning rate of 2e-5 (cosine learning rate scheduler), a batch size of 32, and maximum LLM context length of 2048. All code is written in PyTorch~\cite{paszke2019pytorch}, and we used DeepSpeed~\cite{rasley2020deepspeed} acceleration framework for training. We train our models on two NVIDIA RTX A6000 (48 GB) GPUs, with each training epoch taking approximately 16 hours to complete. During inference, processing of each scanpath (one inference sample) took approximately 1.2 seconds to complete. For inference-time matching for behavior type $\mathbf{D}_{type}$, and stimulus (target category) for Target-Present and Target-Absent search trials, we use the language encoder, MiniLM~\cite{wang2020minilm}, to encode the generated texts and labels in the label-vocabulary and consequently match the generated texts with labels by computing cosine similarities between language embedding of the generated text
and language embedding of each label in the label vocabulary. Finally, we pick the label with the highest cosine similarity. For $\mathbf{D}_{type}$, the label-vocabulary is [\texttt{``Target-Present Visual Search'', ``Target-Absent Visual Search'', ``Object Referral'', ``Visual Question Answering''}]; for target categories in visual search, the label vocabulary is the set of 18 categories in COCO-Search18~\cite{chen2021coco}. Note that for VQA samples, the localization idea unit in a think-aloud transcript is simply a placeholder saying ``The human is answering a question about the image'', since there are no targets to localize. We use the GPT-4o version of GPT-4 to create the think-aloud transcripts and as the evaluation engine under the LLM-as-a-Judge evaluation paradigm. We did not migrate to newer MLLMs to ensure fair comparison with non-LLM baselines. GPT-4o was the strongest and most accessible LLM at the time of this work. Our goal was not simply to maximize performance, but to show that think-aloud instruction tuning is key to adapting MLLMs to human gaze behavior, and that our prompting strategy for LLMs yields effective ground truth for these think-aloud transcripts.

\noindent\textbf{\emph{LLaVA-last}.} \emph{LLaVA-last} is a model that involves predicting the goal based on the last fixation. It is initialized with the weights of Gazette from pre-trained LLaVA-1.5-7B~\cite{liu2024improved} architecture. Then following the same hyperparameters as Gazette, we train the model on gaze decoding with this prompt template: \texttt{\small{Describe the object at co-ordinate (x,y). Use referring expressions if there are multiple \\objects of the same category, else just generate the object category which you can choose from [bottle, bowl, car, chair, clock, cup, fork, keyboard, knife, laptop, microwave, mouse, oven, potted plant, sink, stop sign, toilet, tv].}} This template is also used for inference. Note that mentioning the list of categories helps for LLaVA-last, but not for Gazette, as revealed by our experiments -- perhaps because LLaVA-last is trained on only Object Referral and Target-Present Visual Search, but Gazette is trained on additional Target-Absent Visual Search and VQA, the latter not having any singular target. 

\noindent\textbf{LLaVA-1.5.} This baseline is the frozen, pretrained LLaVA-1.5-7B model~\cite{liu2024improved}. We provide the same \emph{GazeDec} prompt, with one modification -- for visual search scanpaths, we provide the set of COCO-Search18 categories in the prompt, similar to \emph{LLaVA-last}.

\noindent\textbf{GazeGNN.} We implemented \emph{GazeGNN}~\cite{wang2024gazegnn} for gaze target classification. Nishiyasu~\etal~\cite{nishiyasu2024gaze} also implemented GazeGNN but this implementation is not public. Specifically, we replace the final classification layer with two classification layers to do the following tasks: (i) predict behavior type (Target-Present or Target-Absent) and (ii) predict stimulus (one of 18 COCO-Search18 categories). This baseline is not extensible to Object Referral and VQA. All hyperparameters are preserved as designated by the authors of GazeGNN.

\section{Qualitative Analysis}
\label{sec:qual}

In this section, we further qualitatively analyze the model-generated predictions and the attention allocation explanations generated by Gazette. The qualitative examples are in Fig.~\ref{fig:qualitative_vqa}, Fig.~\ref{fig:qualitative_tp}, and Fig.~\ref{fig:qualitative_ta}.

 The crucial contribution of \emph{ThinkAloud} task is underscored by the model behavior for VQA scenarios, as shown in Fig.~\ref{fig:qualitative_vqa}, where complex reasoning patterns are at play. We see that Gazette trained with \emph{ThinkAloud} interprets the scanpath correctly and manages to reconstruct the question from the scanpath to a great degree, even though it asks about ``curtains'' instead of the ``window'' they cover. On the other hand, without \emph{ThinkAloud}, Gazette miserably fails to decode the question corresponding to this scanpath. In Fig.~\ref{fig:qualitative_tp}, we see a scanpath for a Target-Present search for a ``TV''. Owing to fixations next to the chair and close to the floor, Gazette trained w/o \emph{ThinkAloud} instructions erroneously predicts a Target-Absent search for a ``potted plant''. As evidenced by the attention allocation explanation generated by Gazette trained with \emph{ThinkAloud} instructions, Gazette correctly identifies the target ``TV'' that is present in the image. Finally, we show a scanpath of Target-Absent for a ``bowl'' in Fig.~\ref{fig:qualitative_ta}. We see that Gazette trained with \emph{ThinkAloud} instructions identifies the Target-Absent search for a ``bowl'' correctly, instead of mispredicting the ``fork'' distractor object category. We posit that this is because the model finds no fixations to a fork's distractor object in the scene, \ie spoon, but to the ``cup'' which is a distractor object for a ``bowl'', as captured within the generated attention allocation explanation.

\def\subFigSz{\linewidth}
\begin{figure*}[ht!]
\centering
\makebox[\subFigSz][h]{\small{Gazette Predicts: \textbf{VQA} for ``Are there curtains to the right of the chair}}\\
\makebox[\subFigSz][h]{\small{ that is to the right of the lamp?''}}
\makebox[\subFigSz][h]{\small{Gazette w/o ThinkAloud Predicts: \textbf{VQA} for \textbf{``Is there a table to the right of the vase?''}}}
\makebox[\subFigSz][h]{\includegraphics[width=0.7\linewidth]{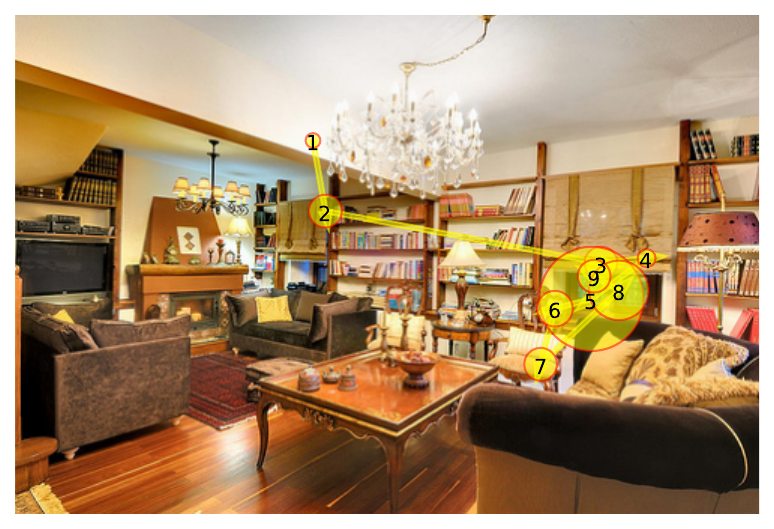} }
    
    \makebox[\subFigSz][h]{
    \begin{minipage}[t]{\linewidth}
      \small{\textbf{Gazette-generated attention allocation explanation}: }\small{Most humans fixated on the chandelier and chain, located centrally in the image, before moving to the curtains and window positioned to the right of the chair, with frequent attention to the light and cabinet near the right edge. }
    \end{minipage}}
    \caption{\textbf{Qualitative Results [1/3].} Comparison of methods on decoding a gaze scanpath corresponding to \textbf{Visual Question Answering} for the question \textbf{``Is the window behind the chair near the pillows?''.} We provide model predictions from full model Gazette, and its variant not trained on \emph{ThinkAloud} instructions. We also provide the attention allocation explanation idea unit in the Gazette-generated think-aloud transcript.}
% \vskip -0.1in
\label{fig:qualitative_vqa}
\end{figure*}

\begin{figure*}[ht!]
\centering
\makebox[\subFigSz][h]{\small{Gazette Predicts: \textbf{Target-Present Search} for \textbf{``TV''}.}}\\
\makebox[\subFigSz][h]{\small{Gazette w/o ThinkAloud Predicts: \textbf{Target-Absent Search} for \textbf{``Potted Plant''}.}}

%tp-000000534121.jpg-tv-9
\makebox[\subFigSz][h]{\includegraphics[width=0.7\linewidth]{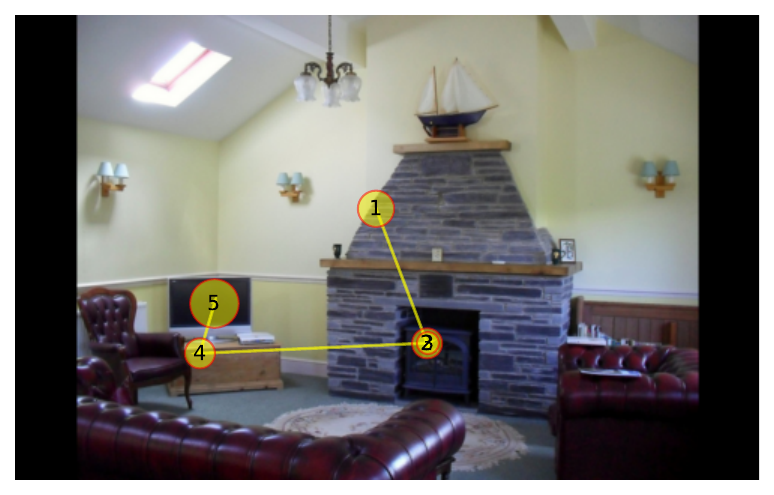} }
\makebox[\subFigSz][h]{\begin{minipage}[t]{\linewidth}
      \small{\textbf{Gazette-generated attention allocation explanation}: }\small{Most humans initially fixate on the wall or fireplace, then shift their gaze towards the vicinity of the TV, often fixating on or near the TV multiple times, with some eventually locating the TV positioned slightly below and to the right of the center of the image.
      }
    \end{minipage}}
    \caption{\textbf{Qualitative Results [2/3]. }Comparison of methods on decoding a gaze scanpath corresponding to \textbf{Target-Present Search} for a \textbf{``TV''}. We provide model predictions from full model Gazette, and its variant not trained on \emph{ThinkAloud} instructions. We also provide the attention allocation explanation idea unit in the Gazette-generated think-aloud transcript.} 
% \vskip -0.1in
\label{fig:qualitative_tp}
\end{figure*}
\begin{figure*}[ht!]
\centering

\makebox[\subFigSz][h]{\small{Gazette Predicts: \textbf{Target-Absent Search} for \textbf{``Bowl''}}}\\
\makebox[\subFigSz][h]{\small{Gazette w/o ThinkAloud Predicts: \textbf{Target-Absent Search} for \textbf{``Fork''}.}}
\makebox[\subFigSz][h]{\includegraphics[width=0.7\linewidth]{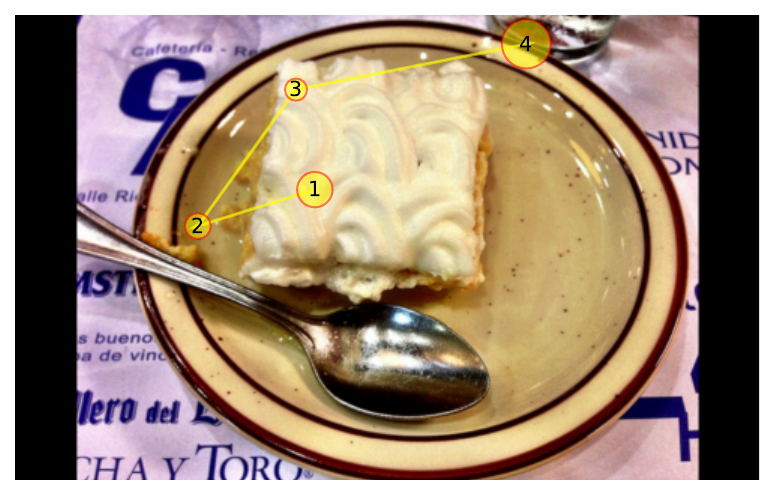} }
    
    \makebox[\subFigSz][h]{
    \begin{minipage}[t]{\linewidth}
      \small{\textbf{Gazette-generated attention allocation explanation}: }\small{Most humans initially fixate on the large cake in the center of the image before shifting their gaze to the cups located on the upper right side, often returning to the cake or intermittently fixating on the dining table. }
    \end{minipage}}
    \caption{\textbf{Qualitative Results [3/3].} Comparison of methods on decoding a gaze scanpath corresponding to \textbf{Target-Absent Search} for a \textbf{``bowl''}. We provide model predictions from full model Gazette, and its variant not trained on \emph{ThinkAloud} instructions. We also provide the attention allocation explanation idea unit in the Gazette-generated think-aloud transcript.} 
% \vskip -0.1in
\label{fig:qualitative_ta}
\end{figure*}
\clearpage
\bibliographystyle{splncs04}
\bibliography{main}

@String(CVPR= {IEEE Conf. Comput. Vis. Pattern Recog.})

@String(AAAI = {AAAI})

@String(VR   = {Vis. Res.})

@String(CVPR  = {CVPR})

@inproceedings{radford2021learning,
  title={Learning transferable visual models from natural language supervision},
  author={Radford, Alec and Kim, Jong Wook and Hallacy, Chris and Ramesh, Aditya and Goh, Gabriel and Agarwal, Sandhini and Sastry, Girish and Askell, Amanda and Mishkin, Pamela and Clark, Jack and others},
  booktitle={International Conference on Machine Learning},
  year={2021}
}

@inproceedings{yu2016modeling,
  title={Modeling context in referring expressions},
  author={Yu, Licheng and Poirson, Patrick and Yang, Shan and Berg, Alexander C and Berg, Tamara L},
  booktitle={European Conference on Computer Vision},
  year={2016}
}

@inproceedings{mondal2023gazeformer,
  title={Gazeformer: Scalable, Effective and Fast Prediction of Goal-Directed Human Attention},
  author={Mondal, Sounak and Yang, Zhibo and Ahn, Seoyoung and Samaras, Dimitris and Zelinsky, Gregory and Hoai, Minh},
  booktitle={Proceedings of the IEEE/CVF Conference on Computer Vision and Pattern Recognition},
  year={2023}
}

@inproceedings{chen2020air,
  title={{AiR}: Attention with reasoning capability},
  author={Chen, Shi and Jiang, Ming and Yang, Jinhui and Zhao, Qi},
  booktitle={European Conference on Computer Vision},
  year={2020}
}

@article{chen2021coco,
  title={{COCO-Search18} fixation dataset for predicting goal-directed attention control},
  author={Chen, Yupei and Yang, Zhibo and Ahn, Seoyoung and Samaras, Dimitris and Hoai, Minh and Zelinsky, Gregory},
  journal={Scientific reports},
  volume={11},
  number={1},
  pages={8776},
  year={2021}}

@article{IttiPAMI98,
	Author = {L. Itti and C. Koch and E. Niebur},
	Journal = {IEEE Transactions on Pattern Analysis and Machine Intelligence},
	Number = {11},
	Pages = {1254--1259},
	Title = {A model of saliency-based visual attention for rapid scene analysis},
	Volume = {20},
	Year = {1998}}

@article{masciocchi2009everyone,
  title={Everyone knows what is interesting: Salient locations which should be fixated},
  author={Masciocchi, Christopher Michael and Mihalas, Stefan and Parkhurst, Derrick and Niebur, Ernst},
  journal={Journal of Vision},
  volume={9},
  number={11},
  pages={25--25},
  year={2009},
  publisher={The Association for Research in Vision and Ophthalmology}
}

@book{yarbus1967eye,
    author = {Yarbus, Alfred L.},
    title = {Eye Movements and Vision},
    year = {1967},
    publisher = {Plenum Press},
    address = {New York},
    translator = {Haigh, Basil}
}

@incollection{henderson2007visual,
  title={Visual saliency does not account for eye movements during visual search in real-world scenes},
  author={Henderson, John M and Brockmole, James R and Castelhano, Monica S and Mack, Michael},
  booktitle={Eye movements},
  pages={537--III},
  year={2007},
  publisher={Elsevier}
}

@article{koehler2014saliency,
  title={What do saliency models predict?},
  author={Koehler, Kathryn and Guo, Fei and Zhang, Sheng and Eckstein, Miguel P},
  journal={Journal of Vision},
  volume={14},
  number={3},
  pages={14--14},
  year={2014},
  publisher={The Association for Research in Vision and Ophthalmology}
}

@article{paszke2019pytorch,
  title={{PyTorch}: An imperative style, high-performance deep learning library},
  author={Paszke, Adam and Gross, Sam and Massa, Francisco and Lerer, Adam and Bradbury, James and Chanan, Gregory and Killeen, Trevor and Lin, Zeming and Gimelshein, Natalia and Antiga, Luca and others},
  journal={Advances in Neural Information Processing Systems},
  year={2019}
}

@inproceedings{yang2022target,
  title={Target-Absent Human Attention},
  author={Yang, Zhibo and Mondal, Sounak and Ahn, Seoyoung and Zelinsky, Gregory and Hoai, Minh and Samaras, Dimitris},
  booktitle={European Conference on Computer Vision},
  year={2022}
}

@inproceedings{chen2022characterizing,
  title={Characterizing Target-Absent Human Attention},
  author={Chen, Yupei and Yang, Zhibo and Chakraborty, Souradeep and Mondal, Sounak and Ahn, Seoyoung and Samaras, Dimitris and Hoai, Minh and Zelinsky, Gregory},
  booktitle={Proceedings of CVPR International Workshop on Gaze Estimation and Prediction in the Wild},
  year={2022}
}

@inproceedings{linMicrosoftCoco2014,
  title = {Microsoft {COCO}: {{Common}} Objects in Context},
  booktitle = {European Conference on Computer Vision},
  author = {Lin, Tsung-Yi and Maire, Michael and Belongie, Serge and Hays, James and Perona, Pietro and Ramanan, Deva and Doll{\'a}r, Piotr and Zitnick, C Lawrence},
  year = {2014},
  pages = {740--755},
  publisher = {{Springer}}
}

@article{zelinskyEyeCan2013,
  title = {Eye Can Read Your Mind: {{Decoding}} Gaze Fixations to Reveal Categorical Search Targets},
  author = {Zelinsky, Gregory J. and Peng, Yifan and Samaras, Dimitris},
  year = {2013},
  journal = {Journal of Vision},
  volume = {13},
  number = {14},
  pages = {10--10},
  issn = {1534-7362}
}

@article{zelinsky2008eye,
  title={Eye can read your mind: Decoding eye movements to reveal the targets of categorical search tasks},
  author={Zelinsky, Gregory and Zhang, Wei and Samaras, Dimitris},
  journal={Journal of Vision},
  volume={8},
  number={6},
  pages={380--380},
  year={2008},
  publisher={The Association for Research in Vision and Ophthalmology}
}

@inproceedings{khokhar2019eye,
  title={Eye-gaze-triggered visual cues to restore attention in educational VR},
  author={Khokhar, A and Yoshimura, A and Borst, CW},
  booktitle={2019 IEEE conference on virtual reality and 3D user interfaces (VR), poster},
  year={2019}
}

@inproceedings{xia2024dream,
  title={{DREAM}: Visual decoding from reversing human visual system},
  author={Xia, Weihao and de Charette, Raoul and Oztireli, Cengiz and Xue, Jing-Hao},
  booktitle={Proceedings of the IEEE/CVF Winter Conference on Applications of Computer Vision},
  year={2024}
}

@inproceedings{takagi2023high,
  title={High-resolution image reconstruction with latent diffusion models from human brain activity},
  author={Takagi, Yu and Nishimoto, Shinji},
  booktitle={Proceedings of the IEEE/CVF Conference on Computer Vision and Pattern Recognition},
  year={2023}
}

@article{defossez2023decoding,
  title={Decoding speech perception from non-invasive brain recordings},
  author={D{\'e}fossez, Alexandre and Caucheteux, Charlotte and Rapin, J{\'e}r{\'e}my and Kabeli, Ori and King, Jean-R{\'e}mi},
  journal={Nature Machine Intelligence},
  volume={5},
  number={10},
  pages={1097--1107},
  year={2023},
  publisher={Nature Publishing Group UK London}
}

@article{benchetrit2023brain,
  title={Brain decoding: toward real-time reconstruction of visual perception},
  author={Benchetrit, Yohann and Banville, Hubert and King, Jean-R{\'e}mi},
  journal={arXiv preprint arXiv:2310.19812},
  year={2023}
}

@article{hollenstein2021decoding,
  title={Decoding {EEG} brain activity for multi-modal natural language processing},
  author={Hollenstein, Nora and Renggli, Cedric and Glaus, Benjamin and Barrett, Maria and Troendle, Marius and Langer, Nicolas and Zhang, Ce},
  journal={Frontiers in Human Neuroscience},
  volume={15},
  pages={659410},
  year={2021},
  publisher={Frontiers Media SA}
}

@article{daly2023neural,
  title={Neural decoding of music from the {EEG}},
  author={Daly, Ian},
  journal={Scientific Reports},
  volume={13},
  number={1},
  pages={624},
  year={2023},
  publisher={Nature Publishing Group UK London}
}

@article{chestek2013hand,
  title={Hand posture classification using electrocorticography signals in the gamma band over human sensorimotor brain areas},
  author={Chestek, Cynthia A and Gilja, Vikash and Blabe, Christine H and Foster, Brett L and Shenoy, Krishna V and Parvizi, Josef and Henderson, Jaimie M},
  journal={Journal of neural engineering},
  volume={10},
  number={2},
  pages={026002},
  year={2013},
  publisher={IOP Publishing}
}

@article{komeiji2024feasibility,
  title={Feasibility of decoding covert speech in {ECoG} with a Transformer trained on overt speech},
  author={Komeiji, Shuji and Mitsuhashi, Takumi and Iimura, Yasushi and Suzuki, Hiroharu and Sugano, Hidenori and Shinoda, Koichi and Tanaka, Toshihisa},
  journal={Scientific Reports},
  volume={14},
  number={1},
  pages={11491},
  year={2024},
  publisher={Nature Publishing Group UK London}
}

@article{bahle2017human,
  title={Human classifier: Observers can deduce task solely from eye movements},
  author={Bahle, Brett and Mills, Mark and Dodd, Michael D},
  journal={Attention, Perception, \& Psychophysics},
  volume={79},
  pages={1415--1425},
  year={2017},
  publisher={Springer}
}

@article{borji2014defending,
  title={Defending {Yarbus}: Eye movements reveal observers' task},
  author={Borji, Ali and Itti, Laurent},
  journal={Journal of Vision},
  volume={14},
  number={3},
  pages={29--29},
  year={2014},
  publisher={The Association for Research in Vision and Ophthalmology}
}

@inproceedings{strohm2021neural,
  title={Neural Photofit: gaze-based mental image reconstruction},
  author={Strohm, Florian and Sood, Ekta and Mayer, Sven and M{\"u}ller, Philipp and B{\^a}ce, Mihai and Bulling, Andreas},
  booktitle={Proceedings of the IEEE/CVF International Conference on Computer Vision},
  year={2021}
}

@inproceedings{strohm2023usable,
  title={Usable and fast interactive mental face reconstruction},
  author={Strohm, Florian and B{\^a}ce, Mihai and Bulling, Andreas},
  booktitle={Proceedings of the 36th Annual ACM Symposium on User Interface Software and Technology},
  year={2023}
}

@inproceedings{strohm2023facial,
  title={Facial composite generation with iterative human feedback},
  author={Strohm, Florian and Sood, Ekta and Thomas, Dominike and B{\^a}ce, Mihai and Bulling, Andreas},
  booktitle={Annual Conference on Neural Information Processing Systems},
  year={2023},
  organization={PMLR}
}

@inproceedings{steil2015discovery,
  title={Discovery of everyday human activities from long-term visual behaviour using topic models},
  author={Steil, Julian and Bulling, Andreas},
  booktitle={Proceedings of the 2015 acm international joint conference on pervasive and ubiquitous computing},
  year={2015}
}

@inproceedings{wang2019mental,
  title={The mental image revealed by gaze tracking},
  author={Wang, Xi and Ley, Andreas and Koch, Sebastian and Lindlbauer, David and Hays, James and Holmqvist, Kenneth and Alexa, Marc},
  booktitle={Proceedings of the 2019 CHI Conference on Human Factors in Computing Systems},
  year={2019}
}

@article{borji2015eyes,
  title={What do eyes reveal about the mind?: Algorithmic inference of search targets from fixations},
  author={Borji, Ali and Lennartz, Andreas and Pomplun, Marc},
  journal={Neurocomputing},
  volume={149},
  pages={788--799},
  year={2015},
  publisher={Elsevier}
}

@article{hu2021ehtask,
  title={{EHTask}: Recognizing user tasks from eye and head movements in immersive virtual reality},
  author={Hu, Zhiming and Bulling, Andreas and Li, Sheng and Wang, Guoping},
  journal={IEEE Transactions on Visualization and Computer Graphics},
  volume={29},
  number={4},
  pages={1992--2004},
  year={2021},
  publisher={IEEE}
}

@article{bulling2010eye,
  title={Eye movement analysis for activity recognition using electrooculography},
  author={Bulling, Andreas and Ward, Jamie A and Gellersen, Hans and Tr{\"o}ster, Gerhard},
  journal={IEEE transactions on pattern analysis and machine intelligence},
  volume={33},
  number={4},
  pages={741--753},
  year={2010},
  publisher={IEEE}
}

@article{bektacs2024gaze,
  title={Gaze-enabled activity recognition for augmented reality feedback},
  author={Bektas, Kenan and Strecker, Jannis and Mayer, Simon and Garcia, Kimberly},
  journal={Computers \& Graphics},
  volume={119},
  pages={103909},
  year={2024},
  publisher={Elsevier}
}

@inproceedings{mondal2025look,
  title={Look Hear: Gaze Prediction for Speech-directed Human Attention},
  author={Mondal, Sounak and Ahn, Seoyoung and Yang, Zhibo and Balasubramanian, Niranjan and Samaras, Dimitris and Zelinsky, Gregory and Hoai, Minh},
  booktitle={European Conference on Computer Vision},
  year={2024}
}

@article{chen2024gazexplain,
  title={{GazeXplain}: Learning to Predict Natural Language Explanations of Visual Scanpaths},
  author={Chen, Xianyu and Jiang, Ming and Zhao, Qi},
  journal={European Conference on Computer Vision},
  year={2024}
}

@inproceedings{sattar2015prediction,
  title={Prediction of search targets from fixations in open-world settings},
  author={Sattar, Hosnieh and Muller, Sabine and Fritz, Mario and Bulling, Andreas},
  booktitle={Proceedings of the IEEE Conference on Computer Vision and Pattern Recognition},
  year={2015}
}

@inproceedings{sattar2017predicting,
  title={Predicting the category and attributes of visual search targets using deep gaze pooling},
  author={Sattar, Hosnieh and Bulling, Andreas and Fritz, Mario},
  booktitle={Proceedings of the IEEE International Conference on Computer Vision Workshops},
  year={2017}
}

@article{sattar2020deep,
  title={Deep gaze pooling: Inferring and visually decoding search intents from human gaze fixations},
  author={Sattar, Hosnieh and Fritz, Mario and Bulling, Andreas},
  journal={Neurocomputing},
  volume={387},
  pages={369--382},
  year={2020},
  publisher={Elsevier}
}

@inproceedings{stauden2018visual,
  title={Visual search target inference using bag of deep visual words},
  author={Stauden, Sven and Barz, Michael and Sonntag, Daniel},
  booktitle={KI 2018: Advances in Artificial Intelligence: 41st German Conference on AI, Berlin, Germany, September 24--28, 2018, Proceedings 41},
  year={2018},
  organization={Springer}
}

@inproceedings{barz2020visual,
  title={Visual search target inference in natural interaction settings with machine learning},
  author={Barz, Michael and Stauden, Sven and Sonntag, Daniel},
  booktitle={ACM Symposium on Eye Tracking Research and Applications},
  year={2020}
}

@inproceedings{nishiyasu2024gaze,
  title={Gaze Scanpath Transformer: Predicting Visual Search Target by Spatiotemporal Semantic Modeling of Gaze Scanpath},
  author={Nishiyasu, Takumi and Sato, Yoichi},
  booktitle={Proceedings of the IEEE/CVF Conference on Computer Vision and Pattern Recognition},
  year={2024}
}

@inproceedings{wang2024gazegnn,
  title={{GazeGNN}: A gaze-guided graph neural network for chest x-ray classification},
  author={Wang, Bin and Pan, Hongyi and Aboah, Armstrong and Zhang, Zheyuan and Keles, Elif and Torigian, Drew and Turkbey, Baris and Krupinski, Elizabeth and Udupa, Jayaram and Bagci, Ulas},
  booktitle={Proceedings of the IEEE/CVF Winter Conference on Applications of Computer Vision},
  year={2024}
}

@article{liaqat2021predicting,
  title={Predicting {ASD} diagnosis in children with synthetic and image-based eye gaze data},
  author={Liaqat, Sidrah and Wu, Chongruo and Duggirala, Prashanth Reddy and Cheung, Sen-ching Samson and Chuah, Chen-Nee and Ozonoff, Sally and Young, Gregory},
  journal={Signal Processing: Image Communication},
  volume={94},
  pages={116198},
  year={2021},
  publisher={Elsevier}
}

@article{li2017implicit,
  title={Implicit intention communication in human--robot interaction through visual behavior studies},
  author={Li, Songpo and Zhang, Xiaoli},
  journal={IEEE Transactions on Human-Machine Systems},
  volume={47},
  number={4},
  pages={437--448},
  year={2017},
  publisher={IEEE}
}

@inproceedings{tawari2014driver,
  title={Where is the driver looking: Analysis of head, eye and iris for robust gaze zone estimation},
  author={Tawari, Ashish and Chen, Kuo Hao and Trivedi, Mohan M},
  booktitle={17th International IEEE conference on intelligent transportation systems (ITSC)},
  year={2014},
  organization={IEEE}
}

@article{buhler2024task,
  title={On Task and in Sync: Examining the Relationship between Gaze Synchrony and Self-Reported Attention During Video Lecture Learning},
  author={B{\"u}hler, Babette and Bozkir, Efe and Deininger, Hannah and Gerjets, Peter and Trautwein, Ulrich and Kasneci, Enkelejda},
  journal={Proceedings of the ACM on Human-Computer Interaction},
  volume={8},
  number={ETRA},
  year={2024},
  publisher={ACM New York, NY, USA}
}

@article{achiam2023gpt,
  title={{GPT-4} technical report},
  author={Achiam, Josh and Adler, Steven and Agarwal, Sandhini and Ahmad, Lama and Akkaya, Ilge and Aleman, Florencia Leoni and Almeida, Diogo and Altenschmidt, Janko and Altman, Sam and Anadkat, Shyamal and others},
  journal={arXiv preprint arXiv:2303.08774},
  year={2023}
}

@article{liu2023visual,
  title={Visual instruction tuning},
  author={Liu, Haotian and Li, Chunyuan and Wu, Qingyang and Lee, Yong Jae},
  journal={Advances in neural information processing systems},
  volume={36},
  pages={34892--34916},
  year={2023}
}

@article{li2024llara,
  title={{LLaRA}: Supercharging robot learning data for vision-language policy},
  author={Li, Xiang and Mata, Cristina and Park, Jongwoo and Kahatapitiya, Kumara and Jang, Yoo Sung and Shang, Jinghuan and Ranasinghe, Kanchana and Burgert, Ryan and Cai, Mu and Lee, Yong Jae and others},
  journal={International Conference on Learning Representations},
  year={2025}
}

@inproceedings{hamza2025llava,
  title={{LLaVA} Needs More Knowledge: Retrieval Augmented Natural Language Generation with Knowledge Graph for Explaining Thoracic Pathologies},
  author={Hamza, Ameer and Ahn, Yong Hyun and Lee, Sungyoung and Kim, Seong Tae and others},
  booktitle={Proceedings of the AAAI Conference on Artificial Intelligence},
  year={2025}
}

@inproceedings{duan2024cityllava,
  title={{CityLLaVA}: Efficient fine-tuning for {VLMs} in city scenario},
  author={Duan, Zhizhao and Cheng, Hao and Xu, Duo and Wu, Xi and Zhang, Xiangxie and Ye, Xi and Xie, Zhen},
  booktitle={Proceedings of the IEEE/CVF Conference on Computer Vision and Pattern Recognition},
  year={2024}
}

@inproceedings{mohbat2024llava,
  title={{LLaVA-Chef}: A multi-modal generative model for food recipes},
  author={Mohbat, Fnu and Zaki, Mohammed J},
  booktitle={Proceedings of the 33rd ACM International Conference on Information and Knowledge Management},
  year={2024}
}

@inproceedings{liu2024improved,
  title     = {Improved Baselines with Visual Instruction Tuning},
  author    = {Liu, Haotian and Li, Chunyuan and Li, Yuheng and Lee, Yong Jae},
  booktitle = {Proceedings of the IEEE/CVF Conference on Computer Vision and Pattern Recognition (CVPR)},
  year      = {2024}
}

@article{driess2023palm,
  title        = {{PaLM-E}: An Embodied Multimodal Language Model},
  author       = {Driess, Danny and Xia, Fei and Sajjadi, Mehdi S. M. and Lynch, Corey and Chowdhery, Aakanksha and Ichter, Brian and Wahid, Ayzaan and Tompson, Jonathan and Vuong, Quan and Yu, Tianhe and Huang, Wenlong and Chebotar, Yevgen and Sermanet, Pierre and Duckworth, Daniel and Levine, Sergey and Vanhoucke, Vincent and Hausman, Karol and Toussaint, Marc and Greff, Klaus and Zeng, Andy and Mordatch, Igor and Florence, Pete},
  journal={arXiv preprint arXiv:2303.03378},
  year={2023}
}

@article{singhal2023clinical,
  title   = {Large language models encode clinical knowledge},
  author  = {Singhal, Karan and Azizi, Sara and Tu, Tong and others},
  journal = {Nature},
  volume  = {620},
  pages   = {297--302},
  year    = {2023}
}

@inproceedings{ranasinghe2024learning,
  title={Learning to localize objects improves spatial reasoning in visual-{LLMs}},
  author={Ranasinghe, Kanchana and Shukla, Satya Narayan and Poursaeed, Omid and Ryoo, Michael S and Lin, Tsung-Yu},
  booktitle={Proceedings of the IEEE/CVF Conference on Computer Vision and Pattern Recognition},
  year={2024}
}

@article{tsang2010eseetrack,
  title={eSeeTrack—visualizing sequential fixation patterns},
  author={Tsang, Hoi Ying and Tory, Melanie and Swindells, Colin},
  journal={IEEE Transactions on Visualization and Computer Graphics},
  volume={16},
  number={6},
  pages={953--962},
  year={2010},
  publisher={IEEE}
}

@article{williams2019changing,
  title={The changing landscape: High-level influences on eye movement guidance in scenes},
  author={Williams, Carrick C and Castelhano, Monica S},
  journal={Vision},
  volume={3},
  number={3},
  pages={33},
  year={2019},
  publisher={MDPI}
}

@article{sabab2022vis,
  title={{Vis-iTrack}: Visual intention through gaze tracking using low-cost webcam},
  author={Sabab, Shahed Anzarus and Kabir, Mohammad Ridwan and Hussain, Sayed Rizban and Mahmud, Hasan and Rubaiyeat, Husne Ara and Hasan, Md Kamrul},
  journal={IEEE Access},
  volume={10},
  pages={70779--70792},
  year={2022},
  publisher={IEEE}
}

@inproceedings{weber2024pattern,
  title={Large Language Models are Pattern Matchers: Editing Semi-Structured and Structured Documents with {ChatGPT}},
  author={Weber, Irene},
  booktitle={AKWI Jahrestagung 2024},
  year={2024}
}

@inproceedings{hudson2019gqa,
  title={{GQA}: A new dataset for real-world visual reasoning and compositional question answering},
  author={Hudson, Drew A and Manning, Christopher D},
  booktitle={Proceedings of the IEEE/CVF conference on computer vision and pattern recognition},
  year={2019}
}

@article{krishna2017visual,
  title={{Visual Genome}: Connecting language and vision using crowdsourced dense image annotations},
  author={Krishna, Ranjay and Zhu, Yuke and Groth, Oliver and Johnson, Justin and Hata, Kenji and Kravitz, Joshua and Chen, Stephanie and Kalantidis, Yannis and Li, Li-Jia and Shamma, David A and others},
  journal={International journal of computer vision},
  volume={123},
  pages={32--73},
  year={2017},
  publisher={Springer}
}

@article{hu2021lora,
  title        = {{LoRA}: Low-Rank Adaptation of Large Language Models},
  author       = {Hu, Edward J. and Shen, Yelong and Wallis, Phillip and Allen-Zhu, Zeyuan
                  and Li, Yuanzhi and Wang, Shean and Wang, Lu and Chen, Weizhu},
  journal      = {arXiv preprint arXiv:2106.09685},
  year         = {2021}
}

@inproceedings{wang2020minilm,
  title     = {{MiniLM}: Deep Self-Attention Distillation for Task-Agnostic Compression of Pre-Trained Transformers},
  author    = {Wenhui Wang and Furu Wei and Li Dong and Hangbo Bao and Nan Yang and Ming Zhou},
  booktitle = {Advances in Neural Information Processing Systems},
  year      = {2020}
}

@inproceedings{papineni-etal-2002-bleu,
  author    = {Kishore Papineni and Salim Roukos and Todd Ward and Wei-Jing Zhu},
  title     = {{BLEU}: a Method for Automatic Evaluation of Machine Translation},
  booktitle = {Proceedings of the 40th Annual Meeting of the Association for Computational Linguistics},
  year      = {2002},
  address   = {Philadelphia, Pennsylvania, USA},
  publisher = {Association for Computational Linguistics}
}

@inproceedings{banerjee-lavie-2005-meteor,
  author    = {Satanjeev Banerjee and Alon Lavie},
  title     = {{METEOR: An automatic metric for MT evaluation with improved correlation with human judgments}},
  booktitle = {Proceedings of the ACL Workshop on Intrinsic and Extrinsic Evaluation Measures for Machine Translation and/or Summarization},
  year      = {2005},
  address   = {Ann Arbor, Michigan},
  publisher = {Association for Computational Linguistics}
}

@inproceedings{lin-2004-rouge,
  author    = {Chin-Yew Lin},
  title     = {{ROUGE}: A Package for Automatic Evaluation of Summaries},
  booktitle = {Text Summarization Branches Out},
  year      = {2004},
  address   = {Barcelona, Spain},
  publisher = {Association for Computational Linguistics},
}

@inproceedings{vedantam-etal-2015-cider,
  author    = {Ramakrishna Vedantam and C. Lawrence Zitnick and Devi Parikh},
  title     = {{CIDEr}: Consensus-based Image Description Evaluation},
  booktitle = {Proceedings of the IEEE Conference on Computer Vision and Pattern Recognition (CVPR)},
  year      = {2015}
}

@article{zheng2023judging,
  title={Judging {LLM-as-a-Judge} with {MT-Bench} and {Chatbot Arena}},
  author={Zheng, Lianmin and Chiang, Wei-Lin and Sheng, Ying and Zhuang, Siyuan and Wu, Zhanghao and Zhuang, Yonghao and Lin, Zi and Li, Zhuohan and Li, Dacheng and Xing, Eric P and Gonzalez, Joseph E and Stoica, Ion},
  journal={arXiv preprint arXiv:2306.05685},
  year={2023}
}

@article{Perfect2020,
  author    = {Perfect, Erin and Hoskin, Elizabeth and Noyek, Samantha and Davies, Claire T.},
  title     = {Outcome Measures and Uptake Barriers When Children and Youth With Complex Disabilities Use Eye Gaze Assistive Technology},
  journal   = {Developmental Neurorehabilitation},
  volume    = {23},
  number    = {3},
  pages     = {145--159},
  year      = {2020},
}

@article{cai2025mergeit,
  title={{MergeIT}: From Selection to Merging for Efficient Instruction Tuning},
  author={Cai, Hongyi and Fu, Yuqian and Fu, Hongming and Zhao, Bo},
  journal={arXiv preprint arXiv:2503.00034},
  year={2025}
}

@article{li2024mosaic,
  title={{Mosaic-IT}: Free Compositional Data Augmentation Improves Instruction Tuning},
  author={Li, Ming and Chen, Pei and Wang, Chenguang and Zhao, Hongyu and Liang, Yijun and Hou, Yupeng and Liu, Fuxiao and Zhou, Tianyi},
  journal={arXiv preprint arXiv:2405.13326},
  year={2024}
}

@book{ericsson1984protocol,
  author    = {Ericsson, K. Anders and Simon, Herbert A.},
  title     = {Protocol Analysis: Verbal Reports as Data},
  publisher = {MIT Press},
  address   = {Cambridge, MA},
  year      = {1984}
}

@book{vansomeren1994think,
  author    = {van Someren, Maarten W. and Barnard, Yvonne F. and Sandberg, Jacobijn A.~C.},
  title     = {The Think Aloud Method: A Practical Guide to Modelling Cognitive Processes},
  publisher = {Academic Press},
  address   = {London},
  year      = {1994},
  isbn      = {9780127142708}
}

@article{fonteyn1993description,
  author  = {Fonteyn, Marsha E. and Kuipers, Brenda and Grobe, Susan J.},
  title   = {A Description of Think Aloud Method and Protocol Analysis},
  journal = {Qualitative Health Research},
  volume  = {3},
  number  = {4},
  pages   = {430--441},
  year    = {1993}
}

@inproceedings{rasley2020deepspeed,
  title={{DeepSpeed}: System optimizations enable training deep learning models with over 100 billion parameters},
  author={Rasley, Jeff and Rajbhandari, Samyam and Ruwase, Olatunji and He, Yuxiong},
  booktitle={Proceedings of the 26th ACM SIGKDD international conference on knowledge discovery \& data mining},
  year={2020}
}

@misc{vicuna2023,
    title = {Vicuna: An Open-Source Chatbot Impressing {GPT-4} with 90\%* {ChatGPT} Quality},
    author = {Chiang, Wei-Lin and Li, Zhuohan and Lin, Zi and Sheng, Ying and Wu, Zhanghao and Zhang, Hao and Zheng, Lianmin and Zhuang, Siyuan and Zhuang, Yonghao and Gonzalez, Joseph E. and Stoica, Ion and Xing, Eric P.},
    month = {March},
    year = {2023}
}

@article{xiong2024llava,
  title={{LLaVA-Critic}: Learning to evaluate multimodal models},
  author={Xiong, Tianyi and Wang, Xiyao and Guo, Dong and Ye, Qinghao and Fan, Haoqi and Gu, Quanquan and Huang, Heng and Li, Chunyuan},
  journal={arXiv preprint arXiv:2410.02712},
  year={2024}
}

@article{mirchandani2023large,
  title={Large language models as general pattern machines},
  author={Mirchandani, Suvir and Xia, Fei and Florence, Pete and Ichter, Brian and Driess, Danny and Arenas, Montserrat Gonzalez and Rao, Kanishka and Sadigh, Dorsa and Zeng, Andy},
  journal={arXiv preprint arXiv:2307.04721},
  year={2023}
}

@article{weber2024large,
  title={Large Language Models are Pattern Matchers: Editing Semi-Structured and Structured Documents with {ChatGPT}},
  author={Weber, Irene},
  journal={arXiv preprint arXiv:2409.07732},
  year={2024}
}

@inproceedings{wang2024vigc,
  title={{VIGC}: Visual instruction generation and correction},
  author={Wang, Bin and Wu, Fan and Han, Xiao and Peng, Jiahui and Zhong, Huaping and Zhang, Pan and Dong, Xiaoyi and Li, Weijia and Li, Wei and Wang, Jiaqi and others},
  booktitle={Proceedings of the AAAI Conference on Artificial Intelligence},
  year={2024}
}

@article{zhang2024llava,
  title={{LLaVA-Video}: Video instruction tuning with synthetic data},
  author={Zhang, Yuanhan and Wu, Jinming and Li, Wei and Li, Bo and Ma, Zejun and Liu, Ziwei and Li, Chunyuan},
  journal={arXiv preprint arXiv:2410.02713},
  year={2024}
}

@InProceedings{Mondal_2025_ICCV,
    author    = {Mondal, Sounak and Sendhilnathan, Naveen and Zhang, Ting and Liu, Yue and Proulx, Michael and Iuzzolino, Michael Louis and Qin, Chuan and Jonker, Tanya R.},
    title     = {Gaze-Language Alignment for Zero-Shot Prediction of Visual Search Targets from Human Gaze Scanpaths},
    booktitle = {Proceedings of the IEEE/CVF International Conference on Computer Vision},
    month     = {October},
    year      = {2025}
}

@article{najemnik2005optimal,
  title={Optimal eye movement strategies in visual search},
  author={Najemnik, Jiri and Geisler, Wilson S},
  journal={Nature},
  volume={434},
  number={7031},
  pages={387--391},
  year={2005},
  publisher={Nature Publishing Group UK London}
}

@article{hayhoe2005eye,
  title={Eye movements in natural behavior},
  author={Hayhoe, Mary and Ballard, Dana},
  journal={Trends in cognitive sciences},
  volume={9},
  number={4},
  pages={188--194},
  year={2005},
  publisher={Elsevier}
}

@inproceedings{xue2026personalized,
  title={Personalized Image Descriptions from Attention Sequences},
  author={Xue, Ruoyu and Le, Hieu and Xu, Jingyi and Mondal, Sounak and Leite, Abe and Zelinsky, Gregory and Hoai, Minh and Samaras, Dimitris},
  booktitle={Proceedings of the IEEE/CVF Conference on Computer Vision and Pattern Recognition},
  year={2026}
}
\end{document}